\documentclass{sn-jnl}

\usepackage[super]{natbib} 
\setcitestyle{comma}

\usepackage[utf8]{inputenc} 
\usepackage{mdframed}
\usepackage{gensymb}%

\usepackage{hyperref}
\hypersetup{
    colorlinks=true,
    linkcolor=blue,
    filecolor=magenta,      
    urlcolor=cyan,
    pdftitle={Causality4FoodSecurity},
    pdfpagemode=FullScreen,
    }
\usepackage{comment}
\usepackage{graphicx}%
\usepackage{multirow}%
\usepackage{amsmath,amssymb,amsfonts}%
\usepackage{amsthm}%
\usepackage{mathrsfs}%
\usepackage[title]{appendix}%
\usepackage{xcolor}%
\usepackage{textcomp}%
\usepackage{manyfoot}%
\usepackage{booktabs}%
\usepackage{algorithm}%
\usepackage{algorithmicx}%
\usepackage{algpseudocode}%
\usepackage{listings}%
\usepackage{longtable}
\usepackage{colortbl} 
\usepackage{subcaption} 
\usepackage{siunitx}

\usepackage{enumitem}
\usepackage{makecell}

\usepackage{xcolor}
\definecolor{nice-green}{HTML}{007849}
\definecolor{nice-blue}{HTML}{0375B4}
\definecolor{nice-orange}{HTML}{CC7722}
\definecolor{nice-red}{HTML}{FF5733}
\usepackage{array}
\newcolumntype{C}[1]{>{\centering\arraybackslash}m{#1}} 
\newcolumntype{L}[1]{>{\raggedright\arraybackslash}m{#1}} 

\usepackage{pifont}
\usepackage{xcolor}
\usepackage{lineno}

\definecolor{midrulegray}{gray}{0.85} 

\newcommand{\lightmidrule}{%
  \noalign{\color{midrulegray}\hrule height 0.3pt}%
}

\newcommand{\lightcmidrules}{%
  \arrayrulecolor{lightgray}%
  \cmidrule(lr){2-5}%
  \arrayrulecolor{black}%
}

\newcommand{\lightallcmidrules}{%
  \arrayrulecolor{lightgray}%
  \cmidrule(lr){1-5}%
  \arrayrulecolor{black}%
}

\newcommand{\lightcmidrulemain}{%
  \arrayrulecolor{lightgray}%
  \cmidrule(lr){2-5}%
  \arrayrulecolor{black}%
}

\newcommand{\lightallcmidrulemain}{%
  \arrayrulecolor{lightgray}%
  \cmidrule(lr){1-5}%
  \arrayrulecolor{black}%
}

\newcommand{\lightcmidrule}{%
  \arrayrulecolor{lightgray}%
  \cmidrule(lr){2-6}%
  \arrayrulecolor{black}%
}

\newcommand{\lightallcmidrule}{%
  \arrayrulecolor{lightgray}%
  \cmidrule(lr){1-6}%
  \arrayrulecolor{black}%
}

\newcommand{\lightcmidrulef}{%
  \arrayrulecolor{lightgray}%
  \cmidrule(lr){2-5}%
  \arrayrulecolor{black}%
}

\newcommand{\lightallcmidrulef}{%
  \arrayrulecolor{lightgray}%
  \cmidrule(lr){1-5}%
  \arrayrulecolor{black}%
}

\newcommand{\lightcmidrulee}{%
  \arrayrulecolor{lightgray}%
  \cmidrule(lr){2-8}%
  \arrayrulecolor{black}%
}

\newcommand{\lightfmidrule}{%
  \arrayrulecolor{lightgray}%
  \cmidrule(lr){2-4}%
  \arrayrulecolor{black}%
}

\newcommand{\lightallcmidrulee}{%
  \arrayrulecolor{lightgray}%
  \cmidrule(lr){1-8}%
  \arrayrulecolor{black}%
}

\definecolor{cblue}{RGB}{164,220,255}
\definecolor{corange}{RGB}{255,187,135}
\definecolor{cgreen}{RGB}{0,139,0} 
\definecolor{psred}{RGB}{246,159,159} 
\definecolor{psblue}{RGB}{142,195,255} 

\definecolor{mutedpurple}{HTML}{9467BD}

\definecolor{qbg}{RGB}{237, 248, 251}
\definecolor{qframe}{RGB}{140, 150, 198}
\definecolor{qtext}{RGB}{140, 150, 198}

\newmdenv[
  backgroundcolor=qbg,
  linecolor=qframe,
  linewidth=1pt,
  roundcorner=6pt,
  innerleftmargin=10pt,
  innerrightmargin=10pt,
  innertopmargin=8pt,
  innerbottommargin=0pt,
  skipabove=10pt,
  skipbelow=10pt
]{questionbox}

\newcommand{\mathbbm}[1]{\text{\usefont{U}{bbm}{m}{n}#1}}

\usepackage{verbatim}

\begin{document}

\UseRawInputEncoding

\title[
Climate Variability Modulates Price-Spike Impacts
]{

Climate Variability Modulates the Impact of Price Spikes on Food Insecurity
}

\author*[1]{Jordi Cerd\`a-Bautista}\email{jordi.cerda@uv.es}
\author[2]{Vasileios Sitokonstantinou}
\author[1]{Homer Durand}
\author[1]{Gherardo Varando}
\author[3]{Michele Ronco}
\author[1]{Gustau Camps-Valls}

\affil[1]{\orgdiv{Image Processing Laboratory}, \orgname{Universitat de Val\`encia}, 
\city{Val\`encia}, 
\country{Spain}}

\affil[2]{\orgdiv{Artificial Intelligence Group}, \orgname{Wageningen University \& Research}, \city{Wageningen}, \country{The Netherlands}}

\affil[3]{\orgdiv{Joint Research Centre}, \orgname{European Commission}, \city{Ispra}, \country{Italy}}

\abstract{
Climate variability influences whether a market disruption escalates into a food crisis, yet broad climate patterns like El Ni\~no, tracked months before they alter hydro-climatic conditions, are still not incorporated as an early-warning component in food-security responses. We address this gap by introducing sensitivity regimes---a stratification of regions by the direction and strength of their vegetation response to the El Ni\~no Southern Oscillation---and using them to estimate how food price spikes affect acute food insecurity across sub-Saharan Africa. Integrating remote sensing, socioeconomic data, and causal machine learning, we find that in regions where ENSO systematically suppresses vegetation, a price spike raises the share of the population at acute risk by 5.4 percentage points in the following month. In regions where vegetation is unaffected by or positively linked to ENSO, the estimated effect is smaller (around 2 percentage points) and statistically insignificant. These results demonstrate that climate context is critical for understanding food security vulnerabilities. Sensitivity regimes can be combined with operational price-spike triggers to stage anticipatory action: the ENSO state flags vulnerable regions months ahead, and a pre-positioned response in those regions to a price spike would avert the largest jump in acute food insecurity.

}

\begin{nolinenumbers}
\maketitle
\end{nolinenumbers}
\clearpage

\begin{nolinenumbers}
\section*{Introduction}
\end{nolinenumbers}

Food insecurity remains a pressing global issue, especially in sub-Saharan Africa, where food systems are increasingly strained by food price spikes, climate variability, and geopolitical instability. According to United Nations estimates, more than 250 million people in Africa are exposed to severe food insecurity, and many additional individuals are at risk as a result of climate change, economic instability, and persistent conflicts~\cite{FAO2023}. Gaining a nuanced understanding of the context-specific drivers of food insecurity is essential to crafting effective policies that bolster resilience and foster sustainable development.

One of the primary drivers of food insecurity is short-run food price spikes, which reduce household purchasing power and can rapidly force changes in consumption and coping strategies~\cite{ivanic2008implications,wodon2010,Barrett2002foodsecurity,headey2008,sen1981}. In operational early-warning systems, food price spikes refer to episodes in which staple prices rise abnormally relative to their expected seasonal pattern. However, the effects of food price spikes are not uniform: they depend on local climatic conditions, agricultural production systems, market structure, and household vulnerability~\cite{kalkuhl2016foodprice,Bellemare2015price,minot2011}. Recent research has highlighted the need to account for these contextual factors when assessing the impact of food price spikes on food security outcomes~\cite{Glauber2020covid,haggblade2017food}. 
When ecosystem and agrosystem conditions for crop production are unfavorable, households often have weaker buffers (e.g., lower agricultural income, reduced own production, and diminished seasonal labor demand). Hence, an abnormal increase in staple prices is more likely to push them into crisis-level coping. When agroclimatic conditions are more favorable, households may be better able to absorb price abnormalities, leading to a smaller marginal impact on acute food insecurity~\cite{WheelerVonBraun2013,brown2008food,myers2017climate,Ansah2021,Gbadegesin2024WPS10999,hlpe2011}. Here, we investigate whether the effects of food price spikes on acute food insecurity can be quantified and how they vary across different climatic contexts.

Climate teleconnections---large-scale anomalies that alter local rainfall, vegetation, and agricultural conditions with a seasonal delay---are a key part of this context. A well-known example is El Ni\~no Southern Oscillation (ENSO): tropical Pacific sea-surface temperature anomalies propagate through atmospheric circulation and systematically influence African hydroclimate and vegetation with seasonal lags of about 1--3 months, varying by region, season, and whether transmission is mediated by rainfall or soil moisture~\cite{McPhaden2006ENSO,brown2008food,Nicholson2014climate,Anyamba2012historical,Sazib2020ENSOAfrica,Philippon2014ENSOAfrica,Anyamba2001NDVIENSO}. This delayed response matters for food security because shifts in harvest prospects, incomes, and coping capacity unfold before price stress materialises. Unlike static agroclimatic zones, ENSO-sensitive regions identify where a remotely observed climate signal is likely to translate into local vulnerability with lead time, providing a natural early-warning layer for anticipatory action before market stress escalates into acute food insecurity.

\begin{figure}[t]
    \centerline{\includegraphics[width=12 cm]{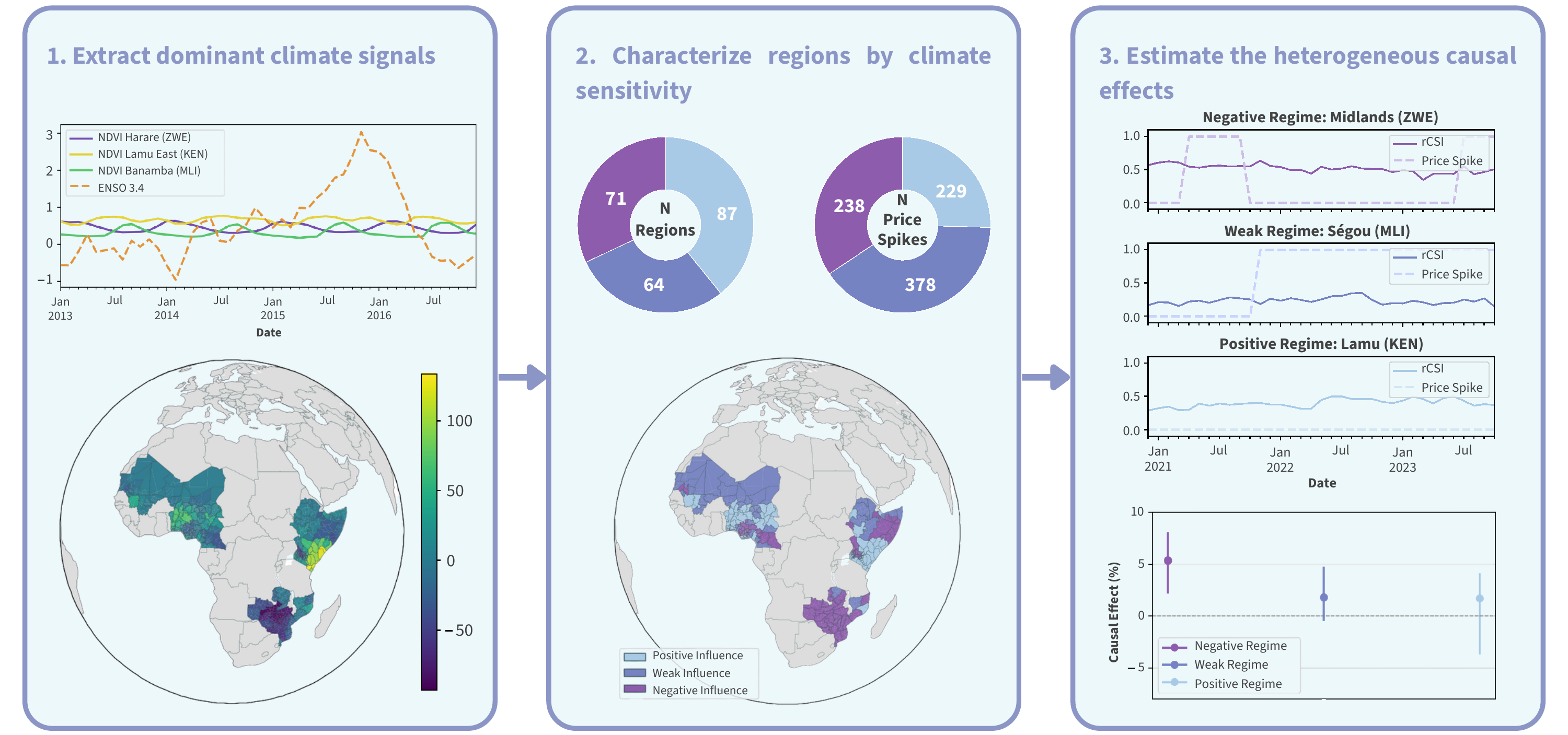}}
    \vspace{0.5cm}
    \caption{\textbf{Proposed analytical pipeline.} The workflow begins by extracting the dominant ENSO-linked vegetation signal, then stratifies regions by sensitivity regimes, and concludes with the estimation of causal effects averaged within each regime. The time-series plots show examples from Kenya, Mali, and Zimbabwe.
    }
    \label{fig:figure_1}
\end{figure}

We present an analytical approach that estimates the effect of food price spikes on acute food insecurity---explicitly conditioned on vegetation sensitivity, as observed through the Normalized Difference Vegetation Index (NDVI)---to ENSO.  Our approach proceeds in three integrated stages, as shown in Figure~\ref{fig:figure_1}. 
First, we extract the dominant ENSO-linked mode of vegetation variability by combining NDVI time series with the Ni\~no~3.4 index and isolating the NDVI component that is temporally driven by ENSO~\cite{Granger1969causality,Shumway2000timeseries,Varando2022grangerpca}. Second, we use this ENSO-linked vegetation signal to define \emph{sensitivity regimes}: areas where ENSO is associated with systematically negative, positive, or weak vegetation responses. These regimes do not summarize average climate conditions; rather, they capture how strongly and in which direction local vegetation responds to ENSO. This differs from static agroclimatic zones, which summarize long-run mean conditions but not dynamic sensitivity to the climate driver. Third, we estimate the causal effect of a price spike within each regime, thereby identifying policy levers that can be toggled before a crisis occurs, which is aligned with early-warning decision needs~\cite{Athey2016causaltrees,Kunzel2019metalearners} (see the \emph{Methods} section for details). By conditioning on these sensitivity regimes, we capture context-specific causal impacts, moving beyond average treatment effects to provide a more granular and nuanced assessment of food security risks~\cite{sitokonstantinou2024causalmachinelearningsustainable,cerdabautista2024assessingcausalimpacthumanitarian,giannarakis2022personalizingsustainableagriculturecausal, benson1994impact,myers2017climate}.

Leveraging remote sensing data, socioeconomic indicators, and causal machine learning, our analysis offers a novel approach to understanding how food price spikes affect food security under climate variability across a wide area of sub-Saharan Africa. 
More broadly, it shows how sensitivity regimes can be combined with price-spike triggers to create an early warning system where pre-positioned responses in vulnerable regions to those price spikes would avert the largest jump in acute food insecurity.

\vspace{0.5cm} 

\begin{nolinenumbers}
\section*{Results}
\end{nolinenumbers}

\begin{nolinenumbers}
\subsection*{\emph{The multivariate food insecurity problem}}
\end{nolinenumbers}

\begin{figure}[t]
    \centerline{\includegraphics[width=12 cm]{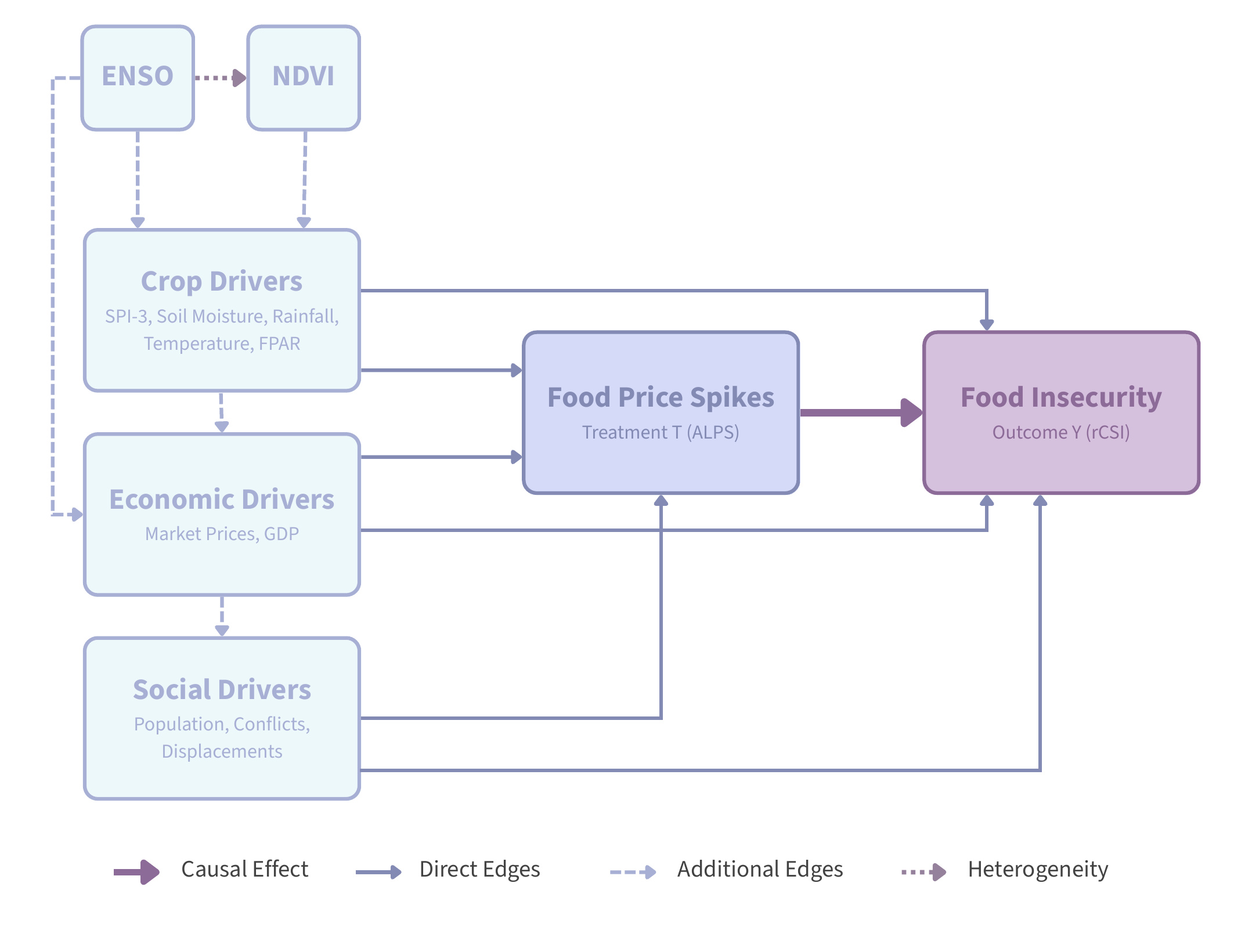}}
    \caption{\textbf{Directed acyclic graph (DAG) representing key determinants of food security in Sub-Saharan Africa and their interactions.} ALPS serves as the treatment, while rCSI represents the outcome of interest. The Food Price Spikes $\rightarrow$ Food Insecurity arrow indicates the causal relationship under investigation. The directed edges represent hypothesized dependencies derived from domain knowledge and the literature.}
    \label{fig:causal_graph}
\end{figure}

Food systems in sub-Saharan Africa are complex, nonlinear, and heavily impacted by climate variability. A causal analysis, therefore, requires both explicit modeling assumptions and a transparent representation of how these drivers may jointly influence exposure and vulnerability. Figure~\ref{fig:causal_graph} shows the directed acyclic graph (DAG) adopted here, grounded in expert knowledge and prior evidence on climate--vegetation dynamics~\cite{Anyamba2018,Ropelewski1987}, conflict and displacement~\cite{IDMC2024GRID}, market stress, and food security outcomes~\cite{FAO2023SOFI,FSIN2024GRFC}. The DAG serves both as a conceptual model of food-system vulnerability and as a guide for selecting adjustment variables in treatment-effect estimation. 

In our study, the treatment of interest is above-normal food price conditions, captured by the Alert for Price Spikes (ALPS) indicator~\cite{WFP2014ALPS}, and the outcome is acute food insecurity measured as the share of the population with a reduced Coping Strategy Index (rCSI)~\cite{WFPCSI2008} in crisis or worse. This outcome is designed to reflect short-term stress in households' ability to access food and is therefore responsive to changes in market conditions.

The DAG doesn't capture every mechanism; it makes explicit which factors drive both treatment and outcome. Its key message is that price stress rarely occurs in isolation. Large-scale climate variability, including ENSO-related teleconnections, influences local agroclimatic conditions such as rainfall, temperature, and soil moisture, which shape vegetation and production conditions and can contribute to market pressure~\cite{Dell2012,Burke2015}. At the same time, conflict and displacement disrupt livelihoods and markets, increasing the likelihood of abnormal prices and reducing households' capacity to cope. Broader economic conditions and demographic structure further shape exposure and vulnerability~\cite{WorldBank2020Pov}.
Our empirical analysis compares observations with similar climate, conflict, displacement, economic, and demographic conditions to estimate how food price spikes are followed by changes in rCSI. This makes our central questions concrete: 
\vspace{0.5cm}

\begin{questionbox}
{\sffamily\bfseries\fontsize{9}{9}\selectfont\color{qtext}
How much do spikes in food prices amplify acute food insecurity in the following month? How does this effect vary across sensitivity regimes?\\
}
\end{questionbox}
The assumed DAG is an abstraction of the system. In reality, food systems involve feedback across time: for example, price spikes can affect displacement and livelihoods, which may later feed back into market conditions. Capturing these dynamics would require an explicitly time-unrolled causal model, whereas the DAG in Figure~\ref{fig:causal_graph} is acyclic by construction and is a necessary assumption for applying causal inference methods. We partially reduce contemporaneous simultaneity by using a one-month lead of the outcome and discuss remaining limitations in the \emph{Discussion} section, and, in \emph{Appendix~A} of the \emph{Supplementary Material}, we report the justifying literature for each causal link. Finally, to quantify any sensitivity to the one-month lead timing choice, we re-estimate the effects across different leads in \emph{Appendix~E} of the \emph{Supplementary Material}.

\vspace{0.5cm}

\begin{nolinenumbers}
\subsection*{\emph{Sensitivity regime stratification for causal impact analysis}}
\end{nolinenumbers}

\begin{figure}[t]
    \hspace*{-0.3cm}
    \centering
    \begin{tabular}{cc}
        \includegraphics[width=1\linewidth]{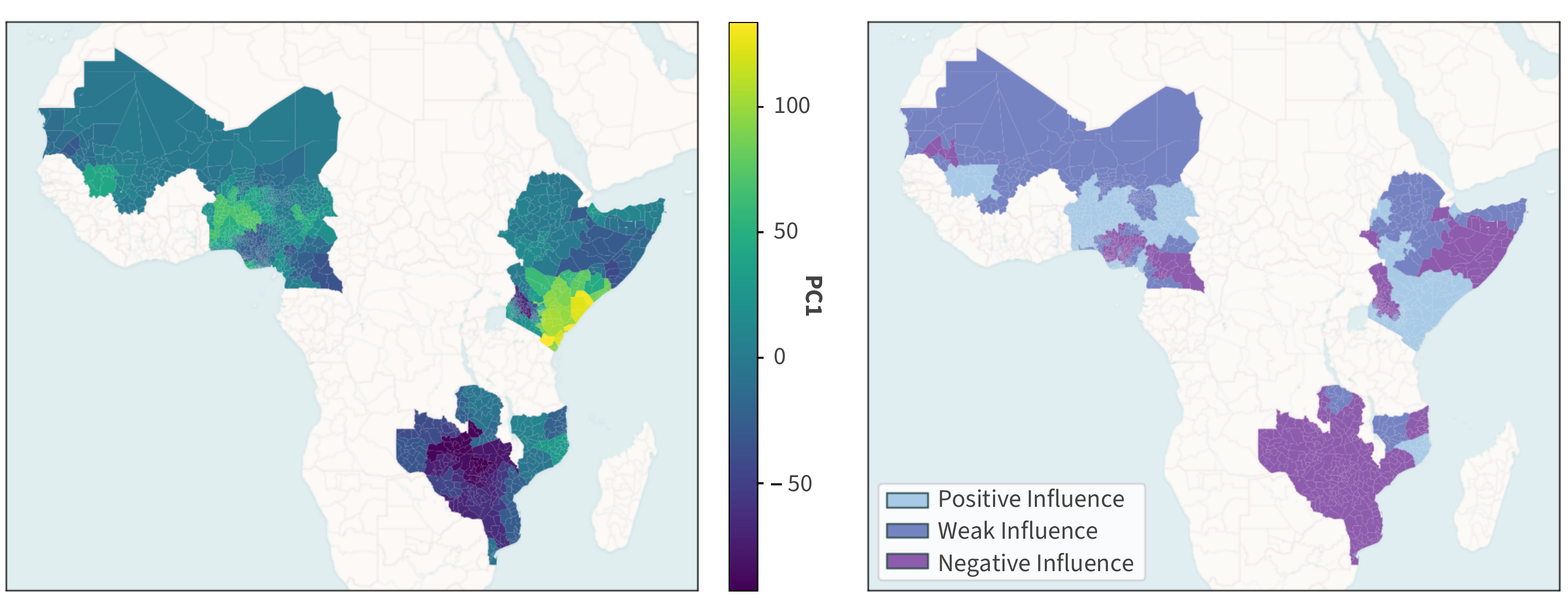}
        
    \end{tabular}
    \caption{\textbf{Geographical distribution of sensitivity regimes based on GPCA stratification.} (Left) First PC of the GPCA analysis on ENSO--NDVI, providing a continuous map of the causal influence. (Right) Regions classified as Positive, Negative, or Weak based on the strength and direction of ENSO--NDVI causal linkages.
    }
    \label{fig:granger_pca}
\end{figure}

Climate variability is a major determinant of food insecurity, and ENSO is among the most important large-scale modes shaping African rainfall and vegetation through teleconnections that operate with seasonal lags~\cite{McPhaden2006ENSO, brown2008food, Nicholson2014climate}. This makes ENSO especially relevant for early warning, since tropical Pacific conditions can signal where vegetation and production are likely to deteriorate months later. NDVI is widely used to monitor these responses, but integrating such high-dimensional, noisy signals into causal effect estimation is challenging~\cite{Anyamba2012historical}. 

We address this with Granger PCA (GPCA)~\cite{Varando2022grangerpca}, a particular instance of Direct Effect Analysis~\cite{durand2025learningcausalresponserepresentations}, which combines NDVI time series with the ENSO index to extract the dominant components of vegetation variability causally driven by ENSO. The output is a set of \emph{sensitivity regimes} — regions where the ENSO anomaly is systematically associated with negative, positive, or weak vegetation responses — that move beyond static agroclimatic zones based on long-run averages.

Figure~\ref{fig:granger_pca} shows the resulting map. \emph{Negative} regions, where ENSO is associated with systematically worse vegetation conditions, broadly align with known drought-prone belts. \emph{Positive} regions correspond to areas with more favorable vegetation conditions, typically zones of more reliable rainfall under the relevant teleconnection states. The \emph{Weak} regime groups regions with a small or mixed ENSO imprint and acts as a buffer between the other two. The regimes thus capture where vegetation is more or less sensitive to a remote driver, not simply where average climates are dry or wet. Implementation details are in the \emph{Methods}; robustness to threshold choices and pixel subsampling is reported in \emph{Appendix~B} of the \emph{Supplementary Material}.

\vspace{0.5cm}

\begin{nolinenumbers}
\subsection*{\emph{Sensitivity regimes modulate the causal effect}}
\end{nolinenumbers}

Our analysis covers 222 administrative level-1 (ADM1) regions observed over 42 months, from September 2020 to February 2024: 71 regions in the Negative regime, 64 in the Weak regime, and 87 in the Positive regime (see the \emph{Methods} subsections \emph{Data} and \emph{Data Preprocessing}). Within this panel, we estimate Conditional Average Treatment Effects (CATEs) of above-normal food price stress on acute food insecurity using two Double Machine Learning estimators: Linear DML and Causal Forest DML. The treatment is defined from the ALPS indicator over a three-month window and contrasts normal food price conditions with above-normal conditions (Stress, Alert, or Crisis). The outcome is the regional monthly prevalence of acute food insecurity, measured as the share of the population with rCSI in crisis or above.

The main result is that the impact of a food price spike depends on climate context (Table~\ref{tab:cate_estimates}). In regions where ENSO tends to suppress vegetation conditions (Negative regime), a price spike is followed by an increase of about 5.4 percentage points (p.p.) in the share of the population experiencing crisis-level coping. In regions where ENSO effects on vegetation are weak or favorable, the estimated increases are much smaller, around 2 p.p., and are not statistically distinguishable from zero. This pattern is consistent with a vulnerability-amplification mechanism. When teleconnected climate conditions repeatedly erode vegetation and production, households enter market-stress episodes with fewer buffers, so the same food price spike translates into a larger increase in acute food insecurity.

To ensure that a single modeling choice does not drive this regime pattern, we compare the DML results with simpler baselines: a difference-in-means estimator, a fixed-effects regression, and a propensity score weighting model. All methods recover the same qualitative ordering, with the largest effects concentrated in the Negative regime (see \emph{Appendix~D} of the \emph{Supplementary Material}).

\begin{table}[t]
\centering
\caption{\textbf{Conditional Average Treatment Effect (CATE) estimates for ALPS impacts on rCSI, stratified by sensitivity regimes.} Confidence intervals are reported at the 95\% level.}
\vspace{0.1cm}
\renewcommand{\arraystretch}{1.}
\begin{tabular}{ccccc}
\bottomrule
\rowcolor[HTML]{ffffff}\textbf{Model} & \textbf{Regime} & \textbf{CATE (p.p.)} & \textbf{C.I.} & \textbf{p-value} \\
\toprule

& Negative & 6.00 & (2.57, 8.73) & 0.020 \\
\lightcmidrulemain
Linear DML & Weak & 1.94 & (-0.47, 4.99) & 0.098 \\
\lightcmidrulemain
& Positive & 2.16 & (-3.92, 4.21) & 0.784 \\
\lightallcmidrulemain

& Negative & 5.35 & (2.16, 8.08) & 0.020 \\
\lightcmidrulemain
Causal Forest DML & Weak & 1.80 & (-0.47, 4.75) & 0.137 \\
\lightcmidrulemain
& Positive & 1.70 & (-3.70, 4.15) & 0.922 \\
\bottomrule
\end{tabular}
\label{tab:cate_estimates}
\end{table}

\begin{figure}[t]
    \hspace*{-0.5cm}
    \centerline{\includegraphics[width=0.9\linewidth]{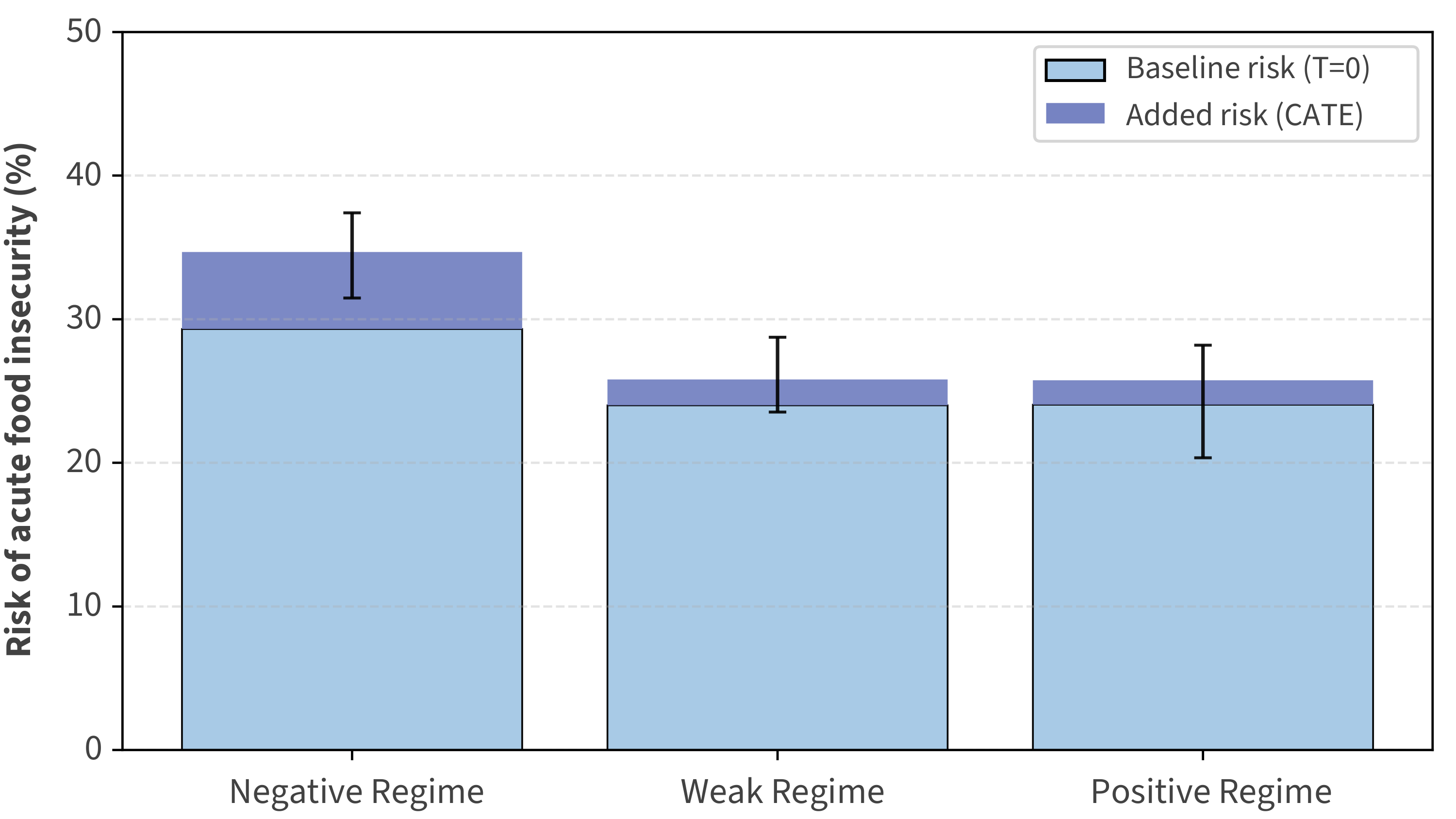}}
    \vspace{0.2cm}
    \caption{\textbf{Baseline risk of acute food insecurity under normal food price conditions and added risk when prices spike.} Estimations provided by the Causal Forest DML CATE estimates stratified for the sensitivity regimes.}

    \label{fig:baseline_added_risk}
\end{figure}

Across outcome leads from 0 to 4 months (\emph{Appendix~E}), the Negative-regime effect remains of similar magnitude and statistically distinguishable from zero, whereas Weak-regime estimates remain small and imprecise. In the Positive regime, estimates are near zero at short horizons and become negative at longer leads; given the observational design and possible differences in seasonality and market integration, we interpret this sign change cautiously as suggestive of different timing or recovery dynamics rather than evidence for a specific mechanism. Pairwise contrasts between regimes follow the expected ordering but are generally not statistically distinguishable from zero (\emph{Appendix~F}). The strongest robust conclusion is therefore that food price spikes have a measurable short-run effect in drought-prone Negative-regime areas, with a directional gradient consistent with the vulnerability-amplification mechanism, but only partially supported by formal pairwise tests.

The regime pattern is most intuitive when viewed alongside the baseline vulnerability under normal food-price conditions (Fig.~\ref{fig:baseline_added_risk}). Regions in the Negative regime already face the highest baseline prevalence of crisis-level coping. A price spike then adds a larger marginal increase on top of an already elevated baseline, pushing more households past critical thresholds. To translate this effect into population terms, we use the median total population across ADM1 regions and the nonlinear estimate for the Negative regime. An ADM1 with a population of approximately 1.32M corresponds to roughly 70{,}600 additional people in crisis or worse in the following month (95\% CI: 28{,}500 to 106{,}700). In weaker or more favorable ENSO--sensitivity contexts, baseline risk is lower, and the marginal effect is smaller and statistically indistinguishable from zero. Taken together, these results suggest that sensitivity regimes provide a practical vulnerability lens for early warning: they help identify where an episode of market stress is most likely to translate into acute food insecurity.

\vspace{0.5cm}

\begin{nolinenumbers}
\subsection*{\emph{Context-specific drivers guide the effect identification}}
\end{nolinenumbers}

\begin{figure}[t]
    \hspace*{-0.6cm}
    \centering
    \includegraphics[width=1.\linewidth]{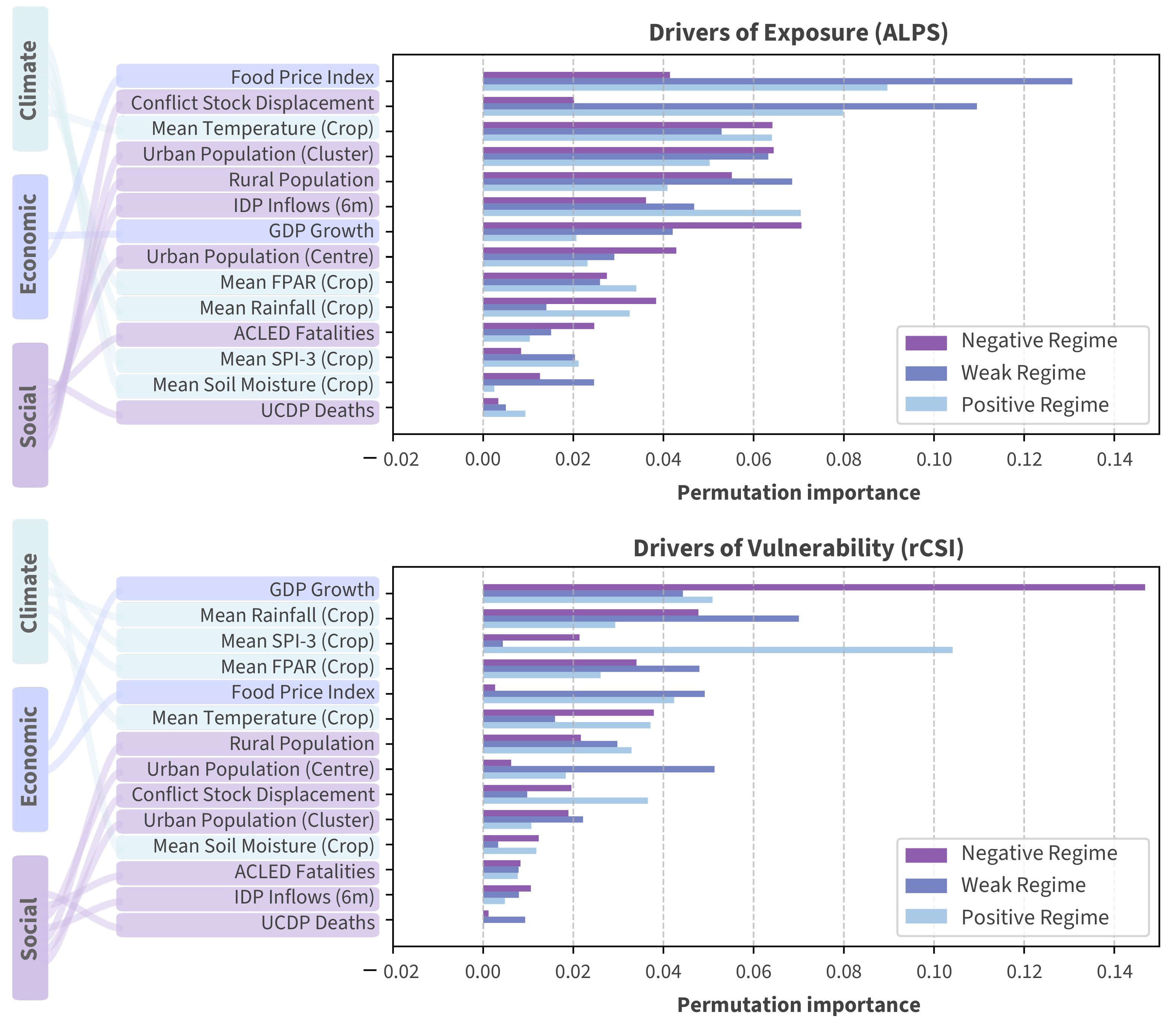}
    \vspace{0.2cm}
    \caption{\textbf{Feature importance of the nuisance models.} Estimations of the treatment (top) and outcome (bottom) for the Causal Forest DML estimator stratified for the sensitivity regimes.}
    \label{fig:feature_importances}
\end{figure}

As part of the DML procedure, we model both the treatment assignment mechanism (exposure to food price spikes) and the outcome regression (vulnerability to acute food insecurity) using machine learning models. Figure~\ref{fig:feature_importances} summarizes which covariates carry the most predictive signal for these two tasks across sensitivity regimes. Covariates fall into three broad domains summarized in the causal graph (Fig.~\ref{fig:causal_graph}): Climate, Economic, and Social. For further information, see the \emph{Methods} section and Table~A1 (\emph{Appendix~A}).

The main message is that the variables that best explain exposure are not the same as those that best explain vulnerability, and their relative importance shifts across ENSO--sensitivity contexts. Exposure is more strongly associated with market and disruption metrics, such as price indices, displacement, and population structure, whereas vulnerability places greater weight on macroeconomic and agroclimatic conditions, as well as conflict/displacement. These feature importances are predictive diagnostics rather than causal attributions.

\vspace{0.5cm}

\begin{nolinenumbers}
\section*{Discussion}
\end{nolinenumbers}

We quantified how food price spikes affect acute food insecurity across distinct sensitivity regimes in sub-Saharan Africa, treating climate not as a control variable but as a source of structured heterogeneity. By defining regimes from ENSO--sensitivity teleconnections rather than static agroclimatic zones, we capture dynamic vulnerability to a remote driver that is monitored months in advance, providing a basis for anticipatory action rather than retrospective adjustment~\cite{hlpe2011}.

The geography of the sensitivity regimes recovers well-known ENSO--rainfall--vegetation patterns: Negative regions concentrate in semi-arid, drought-prone zones of the Horn and South-Eastern Africa, where El Ni\~no conditions typically reduce rainfall and vegetation greenness~\cite{brown2008food, Nicholson2014climate, Anyamba2012historical}, while Positive regions overlap areas where ENSO anomalies coincide with wetter conditions and greener vegetation~\cite{McPhaden2006ENSO, Anyamba2012historical}. The regime-specific effects we estimate are also consistent with documented food-crisis episodes: the 2015--2016 El Ni\~no produced severe drought and elevated food assistance needs across Southern Africa and parts of the Horn~\cite{fao_2016_elNino, winkler2017ENSO_drought}, the areas falling into our Negative regime, where we estimate the largest CATEs and the highest baseline risk. East African regions where El Niño tends to bring above-average rainfall map mainly to the Positive regime, where smaller but positive CATEs are consistent with favorable agroclimate, partially attenuating price-spike impacts even under market stress~\cite{benson1994impact, myers2017climate}.

Across both linear and nonlinear estimators, the Negative regime consistently shows the largest effect, around 5.4 p.p., with Weak and Positive regimes near 2 p.p. (Table~\ref{tab:cate_estimates}). The Negative-regime effect is statistically distinguishable from zero in both estimators and stable across outcome leads, indicating that the clearest, most robust short-run impact of price stress is concentrated in drought-prone, ENSO-sensitive regions. Evidence for between-regime differences is weaker than evidence for a non-zero effect within the Negative regime: point estimates follow the expected ordering (Negative $>$ Weak $\sim$ Positive), but pairwise contrasts are statistically distinguishable from zero only in the linear case for the Negative--versus--Positive comparison (\emph{Appendix~F}). We therefore conclude that food price spikes have a measurable short-run impact in the Negative regime, but the broader gradient across regimes cannot be confirmed through statistical testing. The positive point estimate in the Weak regime informs that, where local production is not strongly ENSO-sensitive, households remain exposed through market channels--dependence on purchased staples, price transmission, and conflict- or displacement-related disruptions~\cite{haggblade2017food, kalkuhl2016foodprice}.

Substantively, regions in the Negative regime exhibit the highest baseline prevalence of crisis-level coping even under normal price conditions, so a price spike adds a larger marginal increase on top of an already elevated baseline, pushing more households across acute-insecurity thresholds. This is consistent with food-system risk architectures in which hazards, exposure, and vulnerability interact rather than operate independently~\cite{FAO2023SOFI, FSIN2024GRFC, Dell2012, Burke2015}: climate does not directly determine food insecurity but shapes the conditions under which a market shock becomes more damaging. 

From a policy perspective, these findings argue against treating the impact of price spikes as a single, uniform number. In the Negative regime, where both baseline risk and marginal effects are high, price-based triggers can be combined with ENSO-informed seasonal outlooks to pre-position assistance, reinforce social protection, and stabilize staple markets before conditions worsen. In the Weak regime, monitoring may need to focus more strongly on conflict, displacement, and market-transmission channels. In the Positive regime, longer-term resilience and agricultural investments may be more relevant, while maintaining attention to conflict-affected or highly urbanized pockets.

Several limitations should temper interpretation. Identification relies on conditional unconfoundedness: informal transfers, local safety nets, targeted aid, and subnational policy responses are not directly observable, and fixed effects can reduce but not eliminate this concern. The Stable Unit Treatment Value Assumption (SUTVA)~\cite{rubin1980randomization, rubin1986comment} is unlikely to hold exactly: prices are transmitted across markets, and conflict and displacement spill over borders, so estimates should be read as within-panel direct-plus-spillover effects, with attenuation toward zero most likely in regimes where market integration is high (one plausible reason non-Negative effects are smaller and less precise). Our climate stratification focuses on ENSO; other remote modes (e.g., the Indian Ocean Dipole) and local extremes (floods, heatwaves) are not explicitly represented. Finally, both ALPS and rCSI are operational indicators with measurement error, so our estimates concern the causal effect of ALPS-defined price stress on harmonized rCSI outcomes, not an abstract true price spike on perfect food security. 

The framework is modular and extensible to other climate drivers, outcomes, or interventions. Substantively, the results highlight a double burden in the Negative sensitivity regime: higher underlying vulnerability and sharper deterioration when prices become abnormal. By using sensitivity regimes as a stratification layer, we show that the short-run food-security impact of price stress is most pronounced in drought-prone regions and smaller elsewhere---a climate-aware perspective for integrating remote climate information into food-security analysis and developing more targeted, anticipatory responses in an era of rising climatic impact, political unrest, and economic volatility.

\vspace{0.5cm}

\begin{nolinenumbers}
\section*{Methods}
\end{nolinenumbers}

\begin{nolinenumbers}
\subsection*{\em Data}
\end{nolinenumbers}

This study uses the Harmonized Food Insecurity Dataset (HFID)~\cite{machefer2025monthlysubnationalharmonizedfood} as the primary data source to model the impact of food price spikes on food insecurity under ENSO--sensitivity variability in Sub-Saharan Africa. The HFID provides sub-national data, integrating multiple sources to capture the complexity of food insecurity dynamics. Among the different food insecurity indicators, the Reduced Coping Strategy Index (rCSI) is a survey-based measure that summarizes the frequency and severity of food-related coping strategies used by households over the previous seven days~\cite{maxwell2008}. In the HFID, rCSI is harmonized and aggregated at the administrative 1 sub-national level; in this study, we focus on the share of the population per region with rCSI at or above the crisis level, which corresponds to households using severe coping strategies and is treated here as an operational threshold for acute food insecurity.

Climate data include monthly measurements of rainfall, temperature, and the fraction of absorbed photosynthetically active radiation (FPAR) as proxies for agricultural productivity, as well as additional drought (SPI) and soil moisture indicators to measure water availability and environmental stress. Socioeconomic data encompass measures such as Gross Domestic Product (GDP) to capture economic performance, population density, conflict data~\cite{ACLED2010,davies2024organized}, and internal displacement~\cite{IDMC2024GRID}.

To quantify abnormal food price conditions, we use the Alert for Price Spikes (ALPS) indicator developed by the World Food Program (WFP)~\cite{WFP2014ALPS, herteux2024}. ALPS is an index that compares observed staple prices to expected seasonal trends and standard deviations, accounting for typical market variability, yielding a continuous measure of price stress. Operationally, WFP uses the threshold $0.25$ to distinguish normal from above-normal price conditions: Normal if ALPS is lower than 0.25, Stress if ALPS is between 0.25 and 1, Alert if ALPS is between 1 and 2, and Crisis if ALPS is higher than 2. We focus on a binary treatment that contrasts normal prices with above-normal price conditions, a policy-relevant distinction for early warning and response.

The dataset used in this work is, therefore, a subset of the HFID that accounts for climatic variables, crop drivers, conflict, displacement, and population proxies, a food price spikes treatment via the ALPS indicator, and a behavioral metric of acute food insecurity that is responsive to short-term stress, via the rCSI. Table A1 in \emph{Appendix~A} of the \emph{Supplementary Material} summarizes all this information. The area of study comprises countries from West Africa (Benin, Cameroon, Mali, Niger, and Nigeria), the Horn of Africa (Ethiopia, Kenya, and Somalia), and South-East Africa (Mozambique, Zambia, and Zimbabwe), and spans from September 2020 until February 2024.

On the other hand, for the sensitivity regime stratification via GPCA analysis, we employ the MODIS Version 6 product MOD13C1 to analyze the impact of ENSO on vegetation~\cite{justice2002overview}. The original global MOD13C1 product contains
NDVI and EVI observations at 16-day temporal resolution and 0.05-degree spatial resolution. 
We consider data from 1st of January, 2001 to 31st of December, 2020, and spatially restrict them to the African continent (\ang{37} S to \ang{38} N, \ang{20} E to \ang{55} W). 
Data have been reprojected into the Equal Area Scalable Earth (EASE) 2.0 25km grid by averaging the original resolution~\cite{ease1, ease2}.\vspace{0.5cm}

\begin{nolinenumbers}
    \subsection*{\em Data preprocessing}
\end{nolinenumbers}

To construct a consistent, analysis-ready panel, we harmonized all data sources to a common spatial and temporal support and derived the treatment, outcome, heterogeneity, and covariate variables for the causal models. We work at the region--month level (ADM1 $\times$ month), indexing each observation by $i=(r,t)$, where $r$ denotes the region and $t$ the month--year. After merging all sources, each row corresponds to a unique region--month with associated food security, price, climate, and conflict information. For each region-month $i=(r,t)$, we construct a confounder vector $W_i$ including: agroclimatic conditions (mean rainfall, soil moisture, temperature, FPAR, and a 3-month mean average of SPI over crop areas), conflict intensity (ACLED fatalities, UCDP deaths),
displacement (IDMC conflict stock displacement, and a 6-month sum of IDPs), macroeconomic conditions (annual GDP growth), market conditions (World Bank food price index), and demography (rural population, urban-center population, and urban-cluster population).

HFID food security indicators, ALPS price indices, climate variables, and conflict data are first aligned to a common monthly time axis over the study period. Climate fields (e.g., rainfall, temperature, soil moisture, NDVI-based indicators) are aggregated from their native grid or station resolution to regional means, while conflict and displacement indicators (e.g., ACLED events, IDP stocks and flows) are summed or averaged within each region and month. All variables are then joined using a harmonized set of ADM1 boundaries, ensuring consistent spatial units across datasets.

To mitigate simultaneity between prices and food security measured in the same month, we define the outcome as next-month acute food insecurity. Let $Y_{r,t}$ denote the population prevalence of households in region $r$ with rCSI in crisis or above in month $t$ (a continuous variable in $[0,1]$). The outcome used in estimation is the one-month lead, $Y_i \equiv Y_{r,t+1}$. As a robustness check on timing, we repeat the CATE analysis with outcome leads $\ell\in\{0,1,2,3,4\}$. Across leads, the Negative-regime effect remains stable and statistically distinguishable from zero, while Weak-regime effects remain smaller and imprecisely estimated; Positive-regime estimates are near zero at short leads and turn negative at longer leads (see \emph{Appendix~E} of the \emph{Supplementary Material}).

The treatment captures abnormal price conditions using ALPS. Let $\text{ALPS}_{r,t}$ denote the (continuous) ALPS index in region $r$ and month $t$. To reduce short-term noise and capture sustained price stress, we use a three-month moving average
\begin{equation}
\overline{\text{ALPS}}_{r,t}^{(3)}=\frac{1}{3}\sum_{k=0}^{2}\text{ALPS}_{r,t-k}.
\end{equation}
We then define the binary treatment as
\begin{equation}
T_i \equiv T_{r,t} =
\begin{cases}
1, & \text{if } \overline{\text{ALPS}}_{r,t}^{(3)} \ge 0.25,\\
0, & \text{otherwise},
\end{cases}
\end{equation}
so that $T_{i}=0$ corresponds to Normal price conditions and $T_{i}=1$ corresponds to above-normal conditions (Stress/Alert/Crisis) under WFP's operational thresholds~\cite{WFP2014ALPS}. This timing choice is also consistent with the temporal scale of the climatic stratification: ENSO-related vegetation responses in Africa are often expressed over seasonal horizons of roughly 1--3 months, with some regions showing lags of about 3 months, so the 3-month ALPS moving average is intended to capture sustained price stress on a similar multi-month scale rather than a point-in-time shock~\cite{Anyamba2001NDVIENSO,Sazib2020ENSOAfrica,FAO2009PriceTransmissionAfrica}.

Each region is assigned to one of three sensitivity regimes derived from the GPCA analysis (Negative, Weak, Positive), as described in the following subsection. This yields a categorical heterogeneity variable $X_i$, which we later use to stratify the CATE estimates.
For each observation, we construct a vector of observed covariates $W_i$ that capture climatic, conflict, displacement, economic, and demographic conditions:
\[
\begin{aligned}
W_i = \big(&\text{rainfall, soil moisture, temperature, FPAR, SPI3, conflict events,} \\
          &\text{displacement, GDP, rural/urban populations}\big).
\end{aligned}
\]
These covariates correspond to the variables used in the adjustment set for the Double Machine Learning models. To absorb time-invariant regional differences (e.g., baseline market access, geography, long-run livelihood structure), we include region fixed effects by adding ADM1 dummy variables to the set of covariates. Otherwise, to absorb global shocks that may affect all geographies equally at a given moment, we include time fixed effects by adding year--month dummy variables, aligned to the $t+1$ time step of the outcome, to the covariate set. Concretely, we augment the adjustment vector with a one-hot encoding of regions and year--month, so that the nuisance models are fit on
\begin{equation}
W_i^{\ast} = \big(W_i,\ \delta_r,\ \delta_{t+1} \big),
\end{equation}
where $\delta_r$ and $\delta_{t+1}$ denote the spatial and temporal fixed effects, respectively.

We remove observations with missing values in the treatment, lead outcome, regime label, or any confounder. Continuous covariates are standardized to have zero mean and unit variance using the training sample to facilitate stable estimation with machine learning models. Categorical variables (including the sensitivity regime and geographic location) are one-hot (dummy) encoded when needed. To avoid regions with insufficient treated/control support, we enforce a minimum within-region support condition, retaining only regions with both treated and control observations over the study window. These steps reduce sensitivity to extreme imbalance and improve the stability of regime-specific estimates. The resulting panel is a monthly region-level dataset where each observation contains $(Y_i, T_i, X_i, W_i^{\ast})$ and is ready for use in the DML pipeline.\vspace{0.5cm}

\begin{nolinenumbers}
    \subsection*{\em Sensitivity regime Stratification via GPCA}
\end{nolinenumbers}

\begin{algorithm}[ht]
\small
\caption{GPCA for sensitivity Regime Classification}
\begin{algorithmic}[1]
\Require NDVI time series $Z_t \in \mathbb{R}^{N \times T}$ for $N$ spatial units over time $T$; ENSO index $E_t$
\Ensure Sensitivity regime label $X_i \in \{ \text{Negative}, \text{Weak}, \text{Positive} \}$ for each unit $r$

\State Construct the data matrix $Z_t = [\text{NDVI}_1^t, \dots, \text{NDVI}_N^t]^T$
\State Compute a low-rank representation via PCA to obtain components $\Phi_k$ and loadings $W$
\State Apply GPCA to obtain rotated components $\widetilde{\Phi}_k$ and select $\widetilde{\Phi}_1$ as the dominant ENSO-linked signal
\State \textbf{Pixel-level weights:} for each pixel $p$, estimate the regression coefficient
\[
\beta_p \;\; \text{from} \;\; \text{NDVI}_p(t) = \alpha_p + \beta_p \widetilde{\Phi}_1(t) + \varepsilon_{p,t},
\]
and sign-correct $\widetilde{\Phi}_1$ so that $\mathrm{corr}(\widetilde{\Phi}_1, E_t)\ge 0$
\State Define a \emph{strong-effect} threshold $t_0 = Q_{0.2}(\mid\beta_p\mid)$ across all pixels
\For{each region $r$}
    \State Compute \textbf{strength} $S_r = \mathrm{median}_{p\in r}(\mid\beta_p\mid)$
    \State Compute \textbf{coverage} $q_r = \Pr_{p\in r}(\mid\beta_p\mid>t_0)$
    \State Compute \textbf{dominance} over strong pixels:
    \[
    D_r = \frac{\sum_{p\in r:\mid\beta_p\mid>t_0} \beta_p a_p}{\sum_{p\in r:\mid\beta_p\mid>t_0} \mid\beta_p\mid a_p},
    \]
    where $a_p=\cos(\mathrm{lat}_p)$ is an area weight.
\EndFor
\State Define a weak threshold $S_{\mathrm{th}} = Q_{0.33}(S_r)$ across regions
\For{each region $r$}
    \If{$S_r < S_{\mathrm{th}}$ \textbf{or} $q_r < 0.2$ \textbf{or} $\mid D_r\mid < 0.2$}
        \State $X_i \leftarrow \text{Weak}$
    \ElsIf{$D_r > 0$}
        \State $X_i \leftarrow \text{Positive}$
    \Else
        \State $X_i \leftarrow \text{Negative}$
    \EndIf
\EndFor
\State \Return $X_i$ for all regions
\end{algorithmic}
\label{alg:granger_pca}
\end{algorithm}

To model ENSO's influence on vegetation, we use satellite-derived NDVI time series across Africa and monthly ENSO indices (Ni\~no 3.4). We capture causal temporal dependencies using Granger-rotated PCA, as shown in Algorithm \ref{alg:granger_pca}. This identifies low-rank NDVI projections that are causally influenced by ENSO.

Using the leading Granger-causal component, we quantify ENSO sensitivity spatially through the regression coefficient $\beta_p$ obtained from $\text{NDVI}_p(t)=\alpha_p+\beta_p \widetilde{\Phi}_1(t)+\varepsilon_{p,t}$ at each pixel $p$. We sign-correct $\widetilde{\Phi}_1$ so that $\mathrm{corr}(\widetilde{\Phi}_1,E_t)\ge 0$, implying $\beta_p>0$ corresponds to higher NDVI under positive ENSO (El Ni\~no-like) conditions. To obtain stable administrative-level regimes, pixels are spatially joined to ADM2 district polygons and aggregated to ADM1 regions. For each ADM1 region $r$, we compute

\begin{itemize}
    \item a robust strength statistic $S_r=\mathrm{median}_{p\in r}(\mid \beta_p\mid )$,
    \item a strong-signal coverage $q_r=\Pr(\mid \beta_p\mid >t_0)$ using $t_0$ defined as the 20th percentile of $\mid \beta_p\mid $ across the area of interest,
    \item a magnitude-weighted sign dominance $D_r=\sum \beta_p a_p / \sum \mid \beta_p\mid  a_p$ over pixels with $\mid \beta_p\mid >t_0$, with $a_p=\cos(\mathrm{lat}_p)$ to approximate equal-area weighting under geographic coordinates.
\end{itemize}
 
Regions are labeled \textit{Weak} if $S_r$ falls below the 33rd percentile across regions, if $q_r<0.2$, or if $\mid D_r\mid <0.2$ (mixed-sign response). Otherwise, regions are labeled \textit{Positive} if $D_r>0$ and \textit{Negative} if $D_r<0$. These ADM1 regime labels are then used as stratification variables in the CATE models. Regime-map robustness to threshold choices and to pixel subsampling is reported in \emph{Appendix~B} of the \emph{Supplementary Material}.\vspace{0.5cm}

\begin{nolinenumbers}
    \subsection*{\em Conditional Average Treatment Effect Estimation}
\end{nolinenumbers}

Estimating the heterogeneous impacts of price spikes on food insecurity is a complex challenge, particularly in regions exposed to significant climate variability. Traditional causal inference methods, such as matching and regression, often struggle to capture the non-linear and high-dimensional relationships present in real-world data. To address these limitations, machine learning approaches like Causal Trees~\cite{Athey2016causaltrees}, Meta-learners~\cite{Kunzel2019metalearners}, and Causal Forests~\cite{wager2017estimationinferenceheterogeneoustreatment} have been developed. These methods provide flexible, nonparametric tools for estimating heterogeneous treatment effects, making them well-suited for analyzing complex, context-specific interventions.

In this work, we adopt the potential outcomes framework~\cite{rubin1974estimating, imbens2015causalInference} to estimate the causal effect of food price spikes on food insecurity. For each unit \( i \), let \( Y_i(1) \) and \( Y_i(0) \) represent the potential outcomes under treatment (i.e., exposure to a price spike) and control (no spike), respectively. The observed outcome is defined as:
\begin{equation}
Y_i = T_i Y_i(1) + (1 - T_i) Y_i(0),
\end{equation}
where \( T_i \in \{0, 1\} \) indicates the binary treatment status, derived from the ALPS indicator. Our target estimand is the Conditional Average Treatment Effect (CATE), defined as:
\begin{equation}
\tau(x) = \mathbb{E}[Y_i(1) - Y_i(0) \mid X_i = x],
\end{equation}
where \( X_i \) is the vector of effect heterogeneity. Treatment effect heterogeneity is examined with respect to the categorical sensitivity regime variable \( X_i \in \{\text{Negative, Weak, Positive}\} \), derived from Granger-rotated PCA (GPCA). We compute stratified average treatment effects within each regime while adjusting for confounders \( W_i \) (climatic, demographic, socioeconomic, and conflict-related features) using flexible machine learning estimators: the Linear DML and the Causal Forest DML. To estimate these effects, we follow a two-stage process: (1) estimate outcome and propensity models using gradient boosting regressors/classifiers; (2) plug the models into CATE estimators to obtain treatment effect estimates and confidence intervals within each sensitivity regime.\vspace{0.5cm}

\begin{nolinenumbers}
    \subsection*{\em Assumptions}
\end{nolinenumbers}

Our causal identification strategy relies on the following assumptions, which enable the estimation of Conditional Average Treatment Effects (CATEs) from observational data.\vspace{0.5cm}

\begin{nolinenumbers}
\subsubsection*{\em Unconfoundedness (Ignorability)}
\end{nolinenumbers}
Formally, this refers to 
\begin{equation}
Y_i(1), Y_i(0) \perp T_i \mid X_i, W_i,
\end{equation}
where we assume that all variables that jointly affect the treatment assignment (price spike exposure) and the outcome (acute food insecurity) are observed in the covariate set \( W_i \), which includes demographic, climate, market, and conflict indicators.\vspace{0.5cm}

\begin{nolinenumbers}
\subsubsection*{\em Positivity (Overlap)}
\end{nolinenumbers}
Positivity refers to the fact that 
\begin{equation}
0 < \mathbb{P}(T_i = 1 \mid X_i, W_i) < 1 \quad,\forall i
\end{equation}
which ensures that for all covariate profiles, there is a non-zero probability of observing both treated and untreated units.\vspace{0.5cm}

\begin{nolinenumbers}
\subsubsection*{\em SUTVA (Stable Unit Treatment Value Assumption)}
\end{nolinenumbers}

We assume no interference between units (e.g., no spillover effects between regions) and that treatment is consistently applied.\vspace{0.5cm}

\begin{nolinenumbers}
\subsubsection*{\em Consistency}
\end{nolinenumbers}

Finally, the observed outcome equals the potential outcome under the treatment actually received:
\begin{equation}
Y_i = Y_i(T_i)    
\end{equation}

\vspace{0.5cm}
The previous assumptions are not directly testable. We aim to make them more plausible by conditioning on a rich set of pre-treatment covariates motivated by the causal graph and domain knowledge, and assessing overlap empirically through propensity score diagnostics and trimming. The DML estimators address high-dimensional, nonlinear confounding adjustment and reduce sensitivity to nuisance-model misspecification through orthogonalization and cross-fitting, but they do not resolve violations due to unobserved confounding or interference.\vspace{0.5cm}

\begin{nolinenumbers}
\subsection*{\em Double Machine Learning for CATE Estimation}
\end{nolinenumbers}

The main effect estimates reported in this work are obtained using Double Machine Learning (DML) estimators, which combine flexible predictive models with Neyman-orthogonal moment conditions to mitigate regularization bias in high-dimensional settings. In our context, DML allows us to estimate EV-regime-specific effects of food price spikes on acute food insecurity while controlling for a rich set of climatic, conflict, displacement, and demographic covariates.\vspace{0.5cm}

\begin{nolinenumbers}
\subsubsection*{\em Setup and notation}
\end{nolinenumbers}

For each observation $i$, let $T_i \in \{0,1\}$ denote the treatment indicator for price spikes (ALPS in Normal vs.\ Stress/Alert/Crisis), $Y_i \in [0,1]$ the prevalence of acute food insecurity (share of population with rCSI in crisis or above), $X_i$ the heterogeneity features (including the sensitivity regime encoded as dummies), and $W_i$ the vector of observed confounders denoted in the \emph{Data Preprocessing} subsection. Under the assumptions in the previous subsection, the target estimand is the Conditional Average Treatment Effect
\begin{equation}
\tau(x) = \mathbb{E}\big[ Y_i(1) - Y_i(0) \,\mid\, X_i = x \big],
\end{equation}
which we summarize by sensitivity regimes $X_i \in \{\text{Negative, Weak, Positive}\}$. We interpret the estimand as the effect of entering an above-normal price-stress state as defined by WFP’s ALPS trigger (ALPS $\geq$ 0.25), relative to remaining in normal conditions, on next-month rCSI prevalence.

\vspace{0.5cm}
\begin{nolinenumbers}
\subsubsection*{\em Orthogonalization and nuisance functions}
\end{nolinenumbers}

Following the DML framework, we decompose the estimation problem into: (i) learning nuisance functions that capture exposure and vulnerability pathways, and (ii) estimating the treatment effect from residualized (orthogonalized) outcomes and treatments. We assume a partially linear model of the form
\begin{equation}
Y_i = m_Y(X_i, W_i^\ast) + T_i \cdot \tau(X_i) + \varepsilon_i,
\qquad \mathbb{E}[\varepsilon_i \mid X_i, W_i^\ast, T_i]=0,
\end{equation}
and a treatment model
\begin{equation}
T_i = m_T(X_i, W_i^\ast) + \nu_i,
\qquad \mathbb{E}[\nu_i \mid X_i, W_i^\ast]=0.
\end{equation}
DML estimates $\tau(\cdot)$ by learning the nuisance functions $m_Y(\cdot)$ and $m_T(\cdot)$ with machine learning models, constructing orthogonalized residuals, and fitting the final-stage effect model on these residuals. We estimate the nuisance functions
\begin{equation}
m_Y(X,W^*) = \mathbb{E}[Y \mid X,W^*], \qquad
m_T(X,W^*) = \mathbb{E}[T \mid X,W^*],
\end{equation}
and, when needed for overlap restriction, the propensity score
\begin{equation}
e(W^*) = \mathbb{P}(T=1 \mid W^*).
\end{equation}
Given estimates $\hat{m}_Y, \hat{m}_T$, we form residuals
\begin{equation}
\tilde{Y}_i = Y_i - \hat{m}_Y(X_i,W_i^*), \qquad
\tilde{T}_i = T_i - \hat{m}_T(X_i,W_i^*),
\end{equation}
and estimate a treatment effect function $\tau(\cdot)$ by fitting the final-stage model using $(\tilde{Y}, \tilde{T}, X_i)$. Because the score function is Neyman-orthogonal, small errors in $\hat{m}_Y$ and $\hat{m}_T$ have only a second-order impact on the final estimates, which is crucial when using flexible machine learning models.\vspace{0.5cm}

\begin{nolinenumbers}
\subsubsection*{\em Cross-fitting}
\end{nolinenumbers}

To reduce overfitting and support the theoretical guarantees of DML while avoiding temporal leakage in panel data, we use $K$-fold cross-fitting with blocked time folds. Concretely, we partition the panel into $K=5$ folds defined by contiguous time blocks (months), and for each fold $k$ we fit nuisance models on the remaining blocks and generate out-of-fold residuals on the held-out block. For each fold $k$, nuisance models are trained on the complement of fold $k$, and residuals $(\tilde{Y}_i, \tilde{T}_i)$ are computed on fold $k$ using only models that were not trained on those observations. The treatment effect model is then fit on the pooled residuals across all folds. This procedure is implemented directly through the \texttt{LinearDML} and \texttt{CausalForestDML} classes from the \texttt{econml}~\cite{econml} package.\vspace{0.5cm}

\begin{nolinenumbers}
\subsubsection*{\em Estimators}
\end{nolinenumbers}

We employ two DML estimators, a Linear DML and a Causal Forest DML. In the LinearDML specification, the final stage assumes a linear treatment effect in a set of basis functions of $X$. Concretely, the estimator fits a penalized linear model of the form
    \begin{equation}
    \tilde{Y}_i = \tilde{T}_i \,\theta^\top \phi(X_i) + \varepsilon_i,
    \end{equation}
where $\phi(X_i)$ includes sensitivity regime dummies, and $\theta$ is estimated via Ridge regression. This yields a parametric, interpretable treatment effect function $\tau(x) = \theta^\top \phi(x)$.

In the CausalForestDML specification, the final stage is a nonparametric causal forest that allows flexible nonlinear interactions among treatment, sensitivity regimes, and covariates. The forest partitions the feature space so that neighboring units with similar $(X,W)$ share similar treatment effect estimates. 

Both estimators treat the outcome as continuous and the treatment as binary. \vspace{0.5cm}

\begin{nolinenumbers}
\subsubsection*{\em Nuisance model selection}
\end{nolinenumbers}

For each nuisance function (outcome regression $m_Y$ and treatment model $m_T$), we perform predictive model selection over a small library of tree-based learners (Random Forest, Gradient Boosting, and XGBoost). We tune hyperparameters via randomized search and evaluate candidates using 5-fold time-block cross-validation, selecting the configuration with the highest validation $R^2$ (for $m_Y$) or ROC--AUC (for $m_T$). The selected learner and hyperparameters are then used within the DML cross-fitting procedure to fit nuisance models on the training folds and generate out-of-fold residuals. This predictive tuning is a practical choice; more targeted selection criteria for doubly robust functionals are an active research topic~\cite{CuiTchetgen2024SelectiveML}. 

In \emph{Appendix~D} of the \emph{Supplementary Material}, we report cross-validated predictive scores (and in-sample diagnostics) for the selected nuisance learners as a check that the covariates capture exposure and vulnerability signal; uncertainty quantification for the causal effects is obtained via the bootstrap described below.
\vspace{0.5cm}

\begin{nolinenumbers}
\subsubsection*{\em Propensity score trimming}
\end{nolinenumbers}

To improve overlap between treated and control observations, we estimate a propensity score $\hat{e}(W^*)$ using logistic regression with standardized covariates~\cite{RosenbaumRubin1983}. To reduce sensitivity to limited overlap and extreme propensity scores, we apply hard trimming, following common-support recommendations in the literature~\cite{CrumpEtAl2009}. In practice, we retain only units satisfying
\begin{equation}
0.2 \leq \hat{e}(W^*) \leq 0.8.
\end{equation}

After trimming to the common-support region, we retain $\approx$ 56\% of observations overall (about 43--64\% across regimes), while preserving non-trivial treated/control support in each regime. The estimators are then fit on this trimmed sample. The relatively large fraction of discarded observations indicates limited overlap in covariate distributions between treated and control units in the full panel, consistent with price-stress episodes being more likely to occur in systematically different contexts (e.g., conflict/displacement and adverse agroclimatic conditions). Consequently, post-trimming estimates should be interpreted as effects for the overlap population in which both treatment states are empirically supported, thereby improving internal validity at the expense of external validity and statistical precision. In \emph{Appendix~C} of the \emph{Supplementary Material}, we provide additional trimming diagnostics (propensity distributions, sample sizes before/after trimming, and standardized mean differences).
\vspace{0.5cm}

\begin{nolinenumbers}
\subsubsection*{\em Sensitivity regime specific effects}
\end{nolinenumbers}

The DML estimators learn an individual-level treatment effect function $\hat{\tau}(x)$. We obtain regime-specific CATEs by averaging these predictions within each sensitivity regime:
\begin{equation}
\widehat{\tau}(x) = \frac{1}{n_x} \sum_{i: X_i = x} \hat{\tau}(X_i),
\end{equation}
where $n_x$ is the number of observations in regime $x$. These are the values reported for Negative, Weak, and Positive regimes in the \emph{Results section}.\vspace{0.5cm}

\begin{nolinenumbers}
\subsubsection*{\em Uncertainty and robustness}
\end{nolinenumbers}

Uncertainty around regime-specific CATE estimates is quantified using a nonparametric bootstrap that accounts for time dependence and common shocks. Specifically, we use a time-block bootstrap: we resample months with replacement from the propensity-score-trimmed panel and retain all regional observations within each selected month, yielding $B = 1000$ bootstrap replicates. For each replicate, we refit the full estimation pipeline (including nuisance models under cross-fitting) and recompute regime-specific CATEs. Percentile-based confidence intervals are computed from the empirical bootstrap distributions, and two-sided $p$-values are computed as
\begin{equation}
p = 2 \min \left\{ \mathbb{P}(\widehat{\tau}^{\ast} \geq 0),\; \mathbb{P}(\widehat{\tau}^{\ast} \leq 0) \right\},
\end{equation}
where $\widehat{\tau}^{\ast}$ denotes the bootstrap replicate of the CATE.

Finally, to probe sensitivity to modeling choices and residual confounding, we run robustness checks: (i) placebo tests where treatment is permuted within regimes, (ii) random common cause (RCC) experiments where a synthetic covariate is added to $W$ and stability of CATEs is assessed, and (iii) random subset removal (RSR), where a fixed fraction of observations is repeatedly dropped and dispersion of CATE estimates is summarized. An additional table with all the robustness metrics can be found in \emph{Appendix~D} of the \emph{Supplementary Material}.\vspace{0.5cm}

\begin{nolinenumbers}
\section*{Data Availability}
\end{nolinenumbers}

The Harmonized Food Insecurity Dataset (HFID) used in this study will be openly available at Zenodo upon publication. 

\vspace{0.5cm}

\begin{nolinenumbers}
\section*{Code Availability}
\end{nolinenumbers}

All code used for the analysis in this paper will be openly available on GitHub at \href{Causal4FS-Prices}{https://github.com/jordicbau/Causal4FS-Prices}. The effect estimation analysis was conducted using the open-source Python packages \href{EconML}{https://econml.azurewebsites.net/} and \href{Scikit-learn}{https://scikit-learn.org/stable/23425}.

\normalsize

\newpage

\begin{nolinenumbers}
\bibliographystyle{unsrtnat}
\bibliography{ref}

\clearpage

\section*{Acknowledgments}

This work was supported by the European Union's Horizon Europe Research and Innovation Program through the \href{ThinkingEarth}{https://thinking-earth.eu/} project (under Grant Agreement number 101130544). GCV research for this study was funded by the European Research Council (ERC) Synergy Grant ``Understanding and Modeling the Earth System with Machine Learning'' (USMILE) under the Horizon 2020 Research and Innovation program (Grant Agreement No. 855187).

\section*{Author Contributions}

J.C-B., V.S., and G.C-V. designed and structured the study. J.C-B. and G.C-V. wrote the initial draft, with contributions and revisions from all co-authors.
H.D. and G.V. provided expertise and implemented GPCA.
J.C-B. led the data harmonization for this specific task. M.R. led the curation, collection, and harmonization of the full HFID. %
G.C-V. supervised the study from inception and secured funding.

\section*{Competing Interests}

The authors declare no competing interests.
\end{nolinenumbers}

\newpage

\linenumbers

\UseRawInputEncoding

\begin{nolinenumbers}

\section*{Supplementary material} 
\end{nolinenumbers}
\begin{appendices}

\begin{nolinenumbers}
\section{Justification of the causal links in the DAG}\label{app:causal_links}
\end{nolinenumbers}

\begin{table}[h]
\centering
\caption{\textbf{Outcome, treatment, and confounders derived from the HFID and used in the causal analysis.}
All variables are aggregated at the ADM1-month level. Confounders are grouped into climate, economic, and social domains.}
\label{tab:variables_summary}
\hspace*{-0.8cm}
\footnotesize
\vspace{0.1cm}
\renewcommand{\arraystretch}{1.15}
\begin{tabular}{C{1.5cm}L{3.5cm}L{5.0cm}C{1.5cm}}
\bottomrule
\rowcolor[HTML]{ffffff}
\textbf{Domain} & \textbf{Variable} & \textbf{Description} & \textbf{Source} \\
\toprule

\textbf{Outcome} &
rCSI &
Reduced Coping Strategy Index crisis prevalence, monthly mean. &
WFP \\
\lightmidrule

\textbf{Treatment} &
ALPS &
Alert for Price Spikes averaged for staple food commodities. &
WFP Prices \\
\lightmidrule

&
Mean Temperature (Crop) &
Spatial mean of the monthly average temperature over crop-covered areas. &
ASAP \\
\lightfmidrule
&
Mean FPAR (Crop) &
Spatial mean of the monthly average fraction of absorbed photosynthetically active radiation (FPAR) over crop-covered areas. &
ASAP \\
\lightfmidrule
\textbf{Climate} &
Mean Rainfall (Crop) &
Spatial mean of the monthly average rainfall over crop-covered areas. &
ASAP \\
\lightfmidrule
&
Mean SPI-3 (Crop) &
Spatial mean of the 3-month Standardized Precipitation Index over crop-covered areas. &
ASAP \\
\lightfmidrule
&
Mean Soil Moisture (Crop) &
Spatial mean of the monthly average combined soil moisture over crop-covered areas. &
ASAP \\
\lightmidrule

\multirow{2}{*}[-2.5ex]{\textbf{Economic}} &
GDP Growth &
Gross Domestic Product annual growth (in \%) of the previous year. &
FAO GDP \\
\lightfmidrule
&
Food Price Index &
Monthly change in international prices of a basket of food commodities. &
World Bank \\
\lightmidrule

\multirow{7}{*}[-12ex]{\textbf{Social} } &
Conflict Stock Displacement &
Number of internally displaced people in stock due to conflict. &
IDMC \\
\lightfmidrule
&
IDP Inflows (6m) &
Number of internally displaced people summed over the last 6 months. &
Rost et al.\ 2025 \\
\lightfmidrule
&
Urban Population (Centre) &
Total population for urban centres (cities). &
GHSL \\
\lightfmidrule
&
Urban Population (Cluster) &
Total population for urban clusters (suburbs/peri-urban). &
GHSL \\
\lightfmidrule
&
Rural Population &
Total rural population. &
GHSL \\
\lightfmidrule
&
ACLED Fatalities &
Number of deaths from conflict events. &
ACLED \\
\lightfmidrule
&
UCDP Deaths &
Number of fatalities from conflict events. &
UCDP \\
\bottomrule
\end{tabular}
\end{table}

\begin{table}[h]
\centering
\caption{Justification of the causal links in the DAG (Part I). For each directed edge we provide a short rationale and example supporting references.}
\label{tab:dag_links_part1}
\footnotesize
\leftskip-.6cm
\renewcommand{\arraystretch}{1.2}
\begin{tabular}{p{1.4cm} p{1.4cm} p{9cm}}
\bottomrule
\rowcolor[HTML]{ffffff}\textbf{From} & \textbf{To} & \textbf{Justification and references} \\
\toprule
ENSO & NDVI, Crop Drivers &
ENSO alters regional rainfall and temperature patterns, which drive vegetation greenness and define crop growing conditions (Anyamba et al., 2018; Ropelewski and Halpert, 1987; FAO et al., 2023). \\
\midrule
ENSO & GDP &
Climate shocks associated with ENSO influence aggregate output and growth, particularly in agriculture-dependent economies (Dell et al., 2012; Burke et al., 2015). \\
\midrule
NDVI & Crop Drivers &
Vegetation condition indices are closely tied to local biophysical crop conditions such as canopy development and water stress (Anyamba et al., 2018; FAO et al., 2023). \\
\midrule
Crop Drivers & GDP &
Weather- and climate-sensitive crop conditions affect yields and agricultural value added, which is a key component of GDP in many low- and middle-income countries (Dell et al., 2012; Burke et al., 2015; FAO et al., 2023). \\
\midrule
Crop Drivers & Conflicts &
Agricultural prod. failures can exacerbate income loss, rural hardship, and competition over resources, recognized as structural and proximate drivers of conflict and instability (Burke et al., 2015; FSIN and Global Network Against Food Crises, 2024). \\
\midrule
Crop Drivers & IDPs &
Agroclimatic shocks contribute to livelihood collapse and displacement, especially in agriculture-dependent settings (FSIN and Global Network Against Food Crises, 2024; IDMC, 2024). \\
\midrule
Crop Drivers & ALPS &
Local weather and crop conditions influence food availability and marketing costs and are core drivers of abnormal local price dynamics, as measured by ALPS (FAO et al., 2023; WFP, 2014). \\
\midrule
Crop Drivers & rCSI &
Adverse crop conditions reduce availability and access to food, increasing the prevalence of crisis-or-worse food consumption and negative coping strategies (FAO et al., 2023; FSIN and Global Network Against Food Crises, 2024). \\
\midrule
GDP & Population &
Economic growth influences fertility, mortality and migration, shaping spatial patterns of rural and urban populations as documented in development and demographic-transition analyses (World Bank, 2020; FAO et al., 2023). \\
\midrule
GDP, Food Prices & IDPs &
Macroeconomic downturns interact with shocks and fragility to increase the likelihood that households resort to displacement as a coping strategy (World Bank, 2020; IDMC, 2024). \\
\midrule
GDP, Food Prices & ALPS &
Macroeconomic conditions affect purchasing power, import costs, and pass-through from global to local food prices, shaping the probability of abnormal price spikes (FAO et al., 2023; FSIN and Global Network Against Food Crises, 2024). \\

\bottomrule
\end{tabular}
\end{table}

\begin{table}[h]
\centering
\caption{Justification of the causal links in the DAG (Part II). For each directed edge we provide a short rationale and example supporting references.}
\label{tab:dag_links_part2}
\footnotesize
\leftskip-.6cm
\renewcommand{\arraystretch}{1.2}
\begin{tabular}{p{1.4cm} p{1.4cm} p{9cm}}
\bottomrule
\rowcolor[HTML]{ffffff}\textbf{From} & \textbf{To} & \textbf{Justification and references} \\
\toprule
GDP, Food Prices & rCSI &
Household income and employment prospects modulated by GDP growth influence the need to adopt negative coping strategies (FAO et al., 2023; World Bank, 2020). \\
\midrule
Conflicts & IDPs &
Violent conflict is the leading global driver of internal displacement (IDMC, 2024; FSIN and Global Network Against Food Crises, 2024). \\
\midrule
Conflicts & ALPS &
Conflict disrupts markets, transport, and production, constraining supply and raising transaction costs (FAO et al., 2023; FSIN and Global Network Against Food Crises, 2024). \\
\midrule
Conflicts & rCSI &
Conflicts reduce incomes, destroy assets and constrain humanitarian access, all of which raise the prevalence of acute food insecurity and severe coping captured by rCSI (FSIN and Global Network Against Food Crises, 2024; FAO et al., 2023). \\
\midrule
IDPs & ALPS &
Inflows of displaced populations can increase local demand, strain markets and infrastructure, and contribute to localised food price spikes (FSIN and Global Network Against Food Crises, 2024; IDMC, 2024). \\
\midrule
IDPs & rCSI &
IDPs face greater exposure to shocks, have constrained access to livelihoods and services, and are consistently among the most food-insecure groups (FSIN and Global Network Against Food Crises, 2024; IDMC, 2024). \\
\midrule
Population & IDPs &
High population concentration in exposed or fragile areas increases the number of people at risk of displacement when conflict or disasters occur (IDMC, 2024; FSIN and Global Network Against Food Crises, 2024). \\
\midrule
Population & ALPS &
Population size and urbanization patterns shape aggregate demand and pressure on food markets and infrastructure (FAO et al., 2023; FSIN and Global Network Against Food Crises, 2024). \\
\midrule
Population & rCSI &
Larger and more rapidly growing, poor populations in fragile settings are associated with higher rates of household food insecurity and negative coping (FAO et al., 2023; World Bank, 2020). \\
\midrule
ALPS & rCSI &
Abnormal food price spikes directly reduce real purchasing power, forcing households to reduce food quantities and quality or adopt harmful coping strategies (WFP, 2014; WFP and CARE, 2008; FSIN and Global Network Against Food Crises, 2024). \\
\bottomrule
\end{tabular}
\end{table}

Our directed acyclic graph (DAG) links large-scale climate variability (ENSO), local biophysical conditions (NDVI, crop-relevant climate), macroeconomic performance (GDP), structural factors (population), violent conflict, internal displacement (IDPs), local food prices (ALPS), and household food security (rCSI). Here, we briefly justify each group of arrows with support from the literature.

\begin{nolinenumbers}
\subsection{ENSO, vegetation, and crop-relevant climate}
\end{nolinenumbers}

El Ni\~no Southern Oscillation (ENSO) is a dominant mode of interannual climate variability and strongly affects rainfall and temperature patterns in many food-insecure regions. Classic work documents systematic ENSO teleconnections in precipitation across the tropics and subtropics, which directly shape soil moisture and drought indices such as SPI. More recent global analyses show that ENSO phases are associated with coherent anomalies in vegetation greenness (NDVI), reflecting impacts on plant growth and crop performance. We therefore include direct arrows from ENSO to NDVI and to the crop-relevant climate variables (rainfall, soil moisture, temperature, FPAR, SPI) grouped as Crop Drivers.

NDVI is both a response to and an integrator of multiple biophysical drivers, especially soil moisture, precipitation and radiation. Empirical studies consistently document strong statistical links between NDVI and concurrent rainfall/soil-moisture anomalies across rain-fed agricultural systems, with NDVI often used as a proxy for realized crop conditions in early-warning systems. We represent this tight coupling with an arrow from NDVI to Crop Drivers, acknowledging that in practice, NDVI and the “Crop Drivers” variables are jointly determined by the same underlying climate processes.\vspace{0.5cm}

\begin{nolinenumbers}
\subsection{Crop conditions, macroeconomy, and food prices}
\end{nolinenumbers}

In low- and middle-income, agriculture-dependent economies, weather shocks to rainfall and temperature have measurable effects on aggregate economic output. A large macroeconomic literature shows that negative temperature and precipitation shocks reduce GDP growth, particularly in poorer and agriculturally intensive countries. Since our Crop Drivers summarize those agro-climatic shocks, we include a directed edge from Crop Drivers to GDP.

Adverse agro-climatic conditions reduce local production and pasture availability, contributing to tighter food markets, higher prices, and lower dietary adequacy. Global and regional assessments from FAO, WFP, and partners consistently highlight weather and climate extremes as major drivers of food price spikes and acute food insecurity. Accordingly, we include arrows from Crop Drivers to ALPS (our indicator of abnormal local food price conditions) and to rCSI (household stress and coping, via their impact on food availability and income opportunities).\vspace{0.5cm}

\begin{nolinenumbers}
\subsection{GDP, population pressure, and food security}
\end{nolinenumbers}

Higher GDP per capita is associated with lower risk of conflict- and disaster-related displacement, in part through improved state capacity, infrastructure, and social protection, while weak or contracting GDP is linked to fragility and higher displacement risk. We capture these long-run relationships with arrows from GDP to Population (through fertility, mortality, and migration) and GDP to IDPs (more prosperous contexts generally experience less forced displacement, all else equal).

Macroeconomic conditions influence both the pass-through of international price shocks and households’ ability to afford food. The State of Food Security and Nutrition in the World reports repeatedly emphasize macroeconomic slowdowns and downturns as core drivers of rising food prices and reduced food access for poor households. We therefore allow GDP to affect both ALPS (through macro-level price dynamics and exchange-rate pass-through) and rCSI (through household incomes and purchasing power).\vspace{0.5cm}

\begin{nolinenumbers}
\subsection{Conflict, displacement and market disruption}
\end{nolinenumbers}

Several studies find that weather-related income shocks can increase the risk of conflict in agriculturally dependent settings, for example, via competition over scarce land and water or reduced state capacity. On this basis, we include an edge from Crop Drivers to Conflicts.

Armed conflict is a leading cause of internal displacement. The Internal Displacement Monitoring Center and related work consistently identify conflict and violence as the dominant source of new and protracted internal displacement in many of the countries covered by our data. This motivates a direct link from Conflicts to IDPs.

Conflict disrupts markets, transport infrastructure, and livelihoods, raising transaction costs, constraining supply, and often causing local food price spikes. At the same time, it limits income-earning opportunities and humanitarian access, deepening acute food insecurity. Global food crisis assessments repeatedly highlight conflict as the primary driver of acute food insecurity episodes, acting through both price and non-price channels. We therefore include outgoing edges from Conflicts to ALPS and to rCSI.\vspace{0.5cm}

\begin{nolinenumbers}
\subsection{Displacement, population, and food security outcomes}
\end{nolinenumbers}

Population density and urbanization patterns condition both the scale of potential displacement and the pressure that displaced and host communities exert on local markets. Higher resident populations and inflows of IDPs into urban and peri-urban areas can strain local food systems and housing, thereby putting upward pressure on prices even when aggregate national production is unchanged. We represent these channels with arrows from Population to IDPs (exposure) and from IDPs to ALPS.

Rapid population growth without commensurate improvements in service delivery and employment opportunities is associated with greater vulnerability to food insecurity in many low-income settings. At the same time, displaced households face particularly high rates of livelihood disruption, asset loss, and food insecurity. Global assessments by WFP and partners consistently report higher rCSI values and worse food security outcomes among IDPs and refugees than among non-displaced populations. We therefore include arrows from both Population and IDPs to rCSI.\vspace{0.5cm}

\begin{nolinenumbers}
\subsection{From price spikes (ALPS) to coping responses (rCSI)}
\end{nolinenumbers}

The reduced Coping Strategy Index (rCSI) is designed to capture the frequency and severity of short-term food-related coping strategies, such as reducing meal portions, skipping meals, or relying on less preferred foods, as households struggle to maintain caloric intake. WFP technical guidelines and validation studies show that rCSI scores rise when households experience deteriorating food access due to income losses or rising staple food prices. Because ALPS (derived from the WFP Price Early Warning Index) explicitly captures abnormal food price conditions, our DAG includes a direct link from ALPS to rCSI, representing the pathway from local price spikes to stress on household food consumption behaviors.\vspace{0.5cm}

\begin{nolinenumbers}
\section{Robustness of the sensitivity regime construction}\label{app:regime_robustness}
\end{nolinenumbers}

Here, we quantify the robustness of the sensitivity regime map derived from the GPCA analysis. The goal is to assess whether the \textit{Positive/Weak/Negative} discretization is sensitive to the regime thresholds used to summarize pixel-level heterogeneity and finite-sample variability in the set of pixels contributing to each administrative unit. All analyses are conducted on the set of countries used in the causal panel (Benin, Cameroon, Mali, Niger, Nigeria, Ethiopia, Kenya, Somalia, Mozambique, Zambia, Zimbabwe).\vspace{0.5cm}

\begin{nolinenumbers}
\subsection{From pixel-level weights to ADM1 regimes}
\end{nolinenumbers}

Let $\beta_p$ denote the pixel-level GPCA weight (sensitivity influence) for pixel $p$, where the GPCA component has been sign-corrected so that $\mathrm{corr}(\widetilde{\Phi}_1,E_t)\ge 0$. Hence, $\beta_p>0$ indicates that positive ENSO anomalies are associated with increased vegetation greenness, and $\beta_p<0$ indicates a negative vegetation response.

To obtain stable administrative-level regimes, pixels are spatially joined to ADM2 polygons and aggregated to ADM1 units. We use three region-level summaries:
(i) strength $S_r = \mathrm{median}_{p\in r}(\mid \beta_p\mid )$, 
(ii) coverage $q_r = \Pr_{p\in r}(\mid \beta_p\mid >t_0)$, and
(iii) dominance over strong pixels,
\begin{equation}
D_r=\frac{\sum_{p\in r:\mid \beta_p\mid >t_0}\beta_p a_p}{\sum_{p\in r:\mid \beta_p\mid >t_0}\mid \beta_p\mid  a_p},
\qquad a_p=\cos(\mathrm{lat}_p),
\end{equation}
where $a_p$ approximates equal-area weighting under geographic coordinates. The threshold $t_0$ is defined as a low quantile of $\mid \beta_p\mid $ across pixels (baseline: 20th percentile), selecting pixels with non-negligible ENSO-related influence.

Regions are labeled \textit{Weak} if any of the following holds:
\begin{equation}
S_r < S_{\mathrm{th}}
\quad \text{or}\quad
q_r < q_{\min}
\quad \text{or}\quad
\mid D_r\mid <d_{\min},
\end{equation}
where $S_{\mathrm{th}}$ is a low quantile of the strength distribution across regions (baseline: 33rd percentile). Otherwise, regions are labeled \textit{Positive} if $D_r>0$ and \textit{Negative} if $D_r<0$. This design intentionally concentrates ambiguity in the \textit{Weak} class (boundary regions), while preserving sign consistency in the \textit{Positive} and \textit{Negative} regimes. \vspace{0.5cm}

\begin{nolinenumbers}
\subsection{Sensitivity to threshold choices}
\end{nolinenumbers}

We evaluate sensitivity to the main regime thresholds by varying:
(i) the strong-pixel quantile used to define $t_0$ (Strong-pixel q),
(ii) the minimum dominance magnitude $d_{\min}$ (Min. Dominance), and
(iii) the minimum strong-signal share $q_{\min}$ (Min. Coverage).
For each configuration, we recompute regime labels for all ADM1 units and compare them to the baseline labeling using: unchanged share (fraction of ADM1 units with identical labels) and Cohen's $\kappa$ (chance-corrected agreement). We also count Positive$\leftrightarrow$Negative sign flips, which are the most consequential changes.

Table~\ref{tab:regime_threshold_sensitivity} summarizes representative settings. Across the tested ranges, agreement with the baseline labeling is high (unchanged share $\ge 0.993$, $\kappa \ge 0.989$), and we observe no direct Positive$\leftrightarrow$Negative sign flips (all changes occur via the \textit{Weak} boundary).

\begin{table}[t]
\centering
\caption{\textbf{Sensitivity of ADM1 regime labels to threshold choices (study countries).}
Agreement metrics compare each configuration to the baseline labeling. Pos$\leftrightarrow$Neg counts direct Positive$\leftrightarrow$Negative sign reversals.}
\label{tab:regime_threshold_sensitivity}
\footnotesize
\hspace*{-1.3cm}
\vspace{0.1cm}
\renewcommand{\arraystretch}{1.2}
\begin{tabular}{rrrrrrrr}
\bottomrule
\rowcolor[HTML]{ffffff}
Strong-pixel q & Min. Dominance & Min. Coverage & Unchanged & $\kappa$ & Pos$\leftrightarrow$Neg & \#Neg & \#Weak/\#Pos \\
\toprule
0.20 & 0.20 & 0.10 & 1.000 & 1.000 & 0 & 158 & 204/203 \\
0.20 & 0.20 & 0.20 & 1.000 & 1.000 & 0 & 158 & 204/203 \\
0.10 & 0.15 & 0.10 & 0.993 & 0.989 & 0 & 159 & 200/206 \\
0.10 & 0.15 & 0.20 & 0.993 & 0.989 & 0 & 159 & 200/206 \\
\bottomrule
\end{tabular}
\end{table}

\vspace{0.5cm}

\begin{nolinenumbers}
\subsection{Stability under pixel subsampling}
\end{nolinenumbers}

Threshold sensitivity addresses parametric choices, but regime assignment can also vary because each ADM1 label is inferred from a finite set of pixels. We therefore run a pixel subsampling stability test: for each ADM1 unit $r$, we repeatedly subsample a fixed fraction of its pixels (keeping their spatial membership), recompute $(S_r,q_r,D_r)$ and the resulting regime label, and estimate the empirical label probabilities
$\Pr(\widehat{X}_r=\ell)$ over $\ell\in\{\text{Negative, Weak, Positive}\}$.

We define the \textbf{stability} of ADM1 unit $r$ as the probability assigned to its baseline label:
\begin{equation}
\mathrm{Stability}_r = \Pr(\widehat{X}_r = X_r^{\mathrm{base}}).
\end{equation}
To isolate the most consequential failure mode, we also compute the Positive$\leftrightarrow$Negative switch probability,
\begin{equation}
\Pr(\text{Pos}\leftrightarrow\text{Neg})_r,
\end{equation}
i.e., the probability that a baseline Positive region is labeled Negative (or vice versa) under subsampling.

In the study countries, baseline regime counts are 71 Negative, 64 Weak, and 87 Positive. Stability is high overall: mean stability is 0.941, and 79.4\% of ADM1 units have stability $\ge 0.9$ (62.2\% have stability $=1.0$). Instability is concentrated in the \textit{Weak} boundary. Crucially, direct sign reversals are essentially absent: the mean Positive$\leftrightarrow$Negative switch probability is $1.39\times 10^{-4}$ (median and 75th percentile equal to 0), with a maximum of 0.02. This indicates that uncertainty is overwhelmingly about whether a region is labeled \textit{Weak} versus a strong-sign regime, not about the sign of the ENSO--Vegetation response.

\begin{table}[t]
\centering
\caption{\textbf{Subsampling stability summary (study countries).}
Stability is the probability of retaining the baseline label when pixels are subsampled. The Pos$\leftrightarrow$Neg switch probability measures direct sign reversals.}
\label{tab:regime_subsampling_summary}
\footnotesize
\vspace{0.1cm}
\renewcommand{\arraystretch}{1.2}
\begin{tabular}{lrr}
\bottomrule
\rowcolor[HTML]{ffffff} Metric & Value & Notes \\
\toprule
\# ADM1 units & 222 & Study countries only \\
Baseline counts (Neg/Weak/Pos) & 71 / 64 / 87 &  \\
Mean stability & 0.941 & $\Pr(\widehat{X}_r = X_r^{\mathrm{base}})$ \\
Share stability $\ge 0.9$ & 0.794 &  \\
Share stability $=1.0$ & 0.622 &  \\
Mean Pos$\leftrightarrow$Neg switch prob. & $1.39\times 10^{-4}$ & Median = 0, 75th pct = 0 \\
Max Pos$\leftrightarrow$Neg switch prob. & 0.020 & Worst-case ADM1 \\
\bottomrule
\end{tabular}
\end{table}

\vspace{0.5cm}

\begin{nolinenumbers}
\subsection{Most unstable units (illustrative)}
\end{nolinenumbers}

To illustrate where instability arises, Table~\ref{tab:regime_low_stability} lists ADM1 units with the lowest stability in the subsampling test. In all cases, probability mass concentrates between \textit{Weak} and one strong-sign regime, whereas sign flips remain negligible.

\begin{table}[t]
\centering
\caption{\textbf{Lowest-stability ADM1 units in the subsampling test (study countries).}
Columns show baseline label, stability, and empirical label probabilities under subsampling.}
\label{tab:regime_low_stability}
\footnotesize
\vspace{0.1cm}
\renewcommand{\arraystretch}{1.2}
\begin{tabular}{llllrrrr}
\bottomrule
\rowcolor[HTML]{ffffff}
GID\_0 & Country & ADM1 & Baseline & Stability & $P(\text{Neg})$ & $P(\text{Weak})$ & $P(\text{Pos})$ \\
\toprule
MOZ & Mozambique & Maputo & Negative & 0.575 & 0.575 & 0.425 & 0.000 \\
NGA & Nigeria & Sokoto & Weak & 0.585 & 0.000 & 0.585 & 0.415 \\
KEN & Kenya & Kericho & Negative & 0.585 & 0.585 & 0.380 & 0.035 \\
NGA & Nigeria & Nasarawa & Weak & 0.590 & 0.410 & 0.590 & 0.000 \\
NGA & Nigeria & Rivers & Positive & 0.600 & 0.005 & 0.395 & 0.600 \\
KEN & Kenya & Turkana & Negative & 0.615 & 0.615 & 0.385 & 0.000 \\
KEN & Kenya & Bungoma & Negative & 0.640 & 0.640 & 0.360 & 0.000 \\
KEN & Kenya & Kiambu & Weak & 0.655 & 0.330 & 0.655 & 0.000 \\
\bottomrule
\end{tabular}
\end{table}

\vspace{0.5cm}

\begin{nolinenumbers}
\section{Propensity Score analysis}\label{app:ps_score}
\end{nolinenumbers}

\begin{nolinenumbers}
\subsection{Calculation}
\end{nolinenumbers}

\begin{figure}[t]
\centering
\includegraphics[width=1\textwidth]{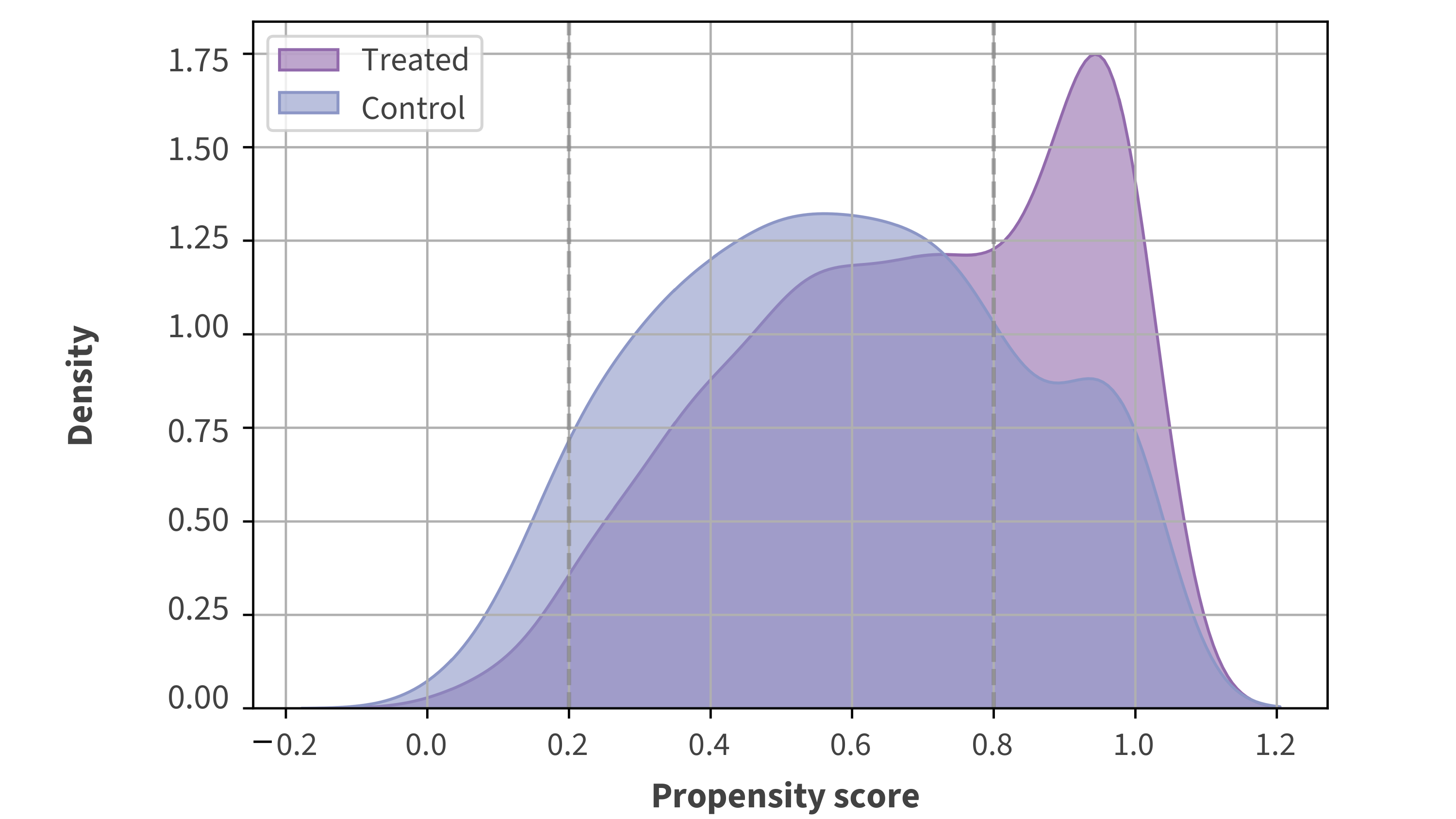}
\vspace{0.2cm}
\caption{\textbf{Kernel density estimates of the propensity scores.} 
The density distributions illustrate the propensity score distributions for the treated and control groups. The vertical dashed lines indicate the lower and upper cutoffs for the propensity-score trimming.}
\label{fig:propensity_scores}
\end{figure}

\begin{table}[t]
\centering
\caption{\textbf{Sample sizes before and after propensity-score trimming, overall and by sensitivity regime.}
We report the number of control ($T{=}0$) and treated ($T{=}1$) observations before trimming and after restricting to the common-support region
$0.2\le \hat e(W_i) \le 0.8$.}
\label{tab:ps_counts_regime}
\footnotesize
\vspace{0.1cm}
\renewcommand{\arraystretch}{1.2}
\begin{tabular}{lrrrr}
\bottomrule
\rowcolor[HTML]{ffffff}
\textbf{Regime} &
\multicolumn{2}{c}{\textbf{Before trimming}} &
\multicolumn{2}{c}{\textbf{After trimming}} \\
\rowcolor[HTML]{ffffff}
& \textbf{Control} & \textbf{Treated} & \textbf{Control} & \textbf{Treated} \\
\toprule
Overall  & 829 & 845 & 483 & 455 \\
\lightmidrule
Negative & 318  & 238 & 227  & 130 \\
Weak     & 271  & 378 & 158  & 220 \\
Positive & 240  & 229 & 98  & 105 \\
\bottomrule
\end{tabular}
\end{table}

In the main analysis, propensity scores are used to (i) assess covariate overlap between treated and control observations and (ii) trim observations with extreme treatment probabilities prior to causal effect estimation. For each region--month observation $i$, we define the propensity score as
\begin{equation}
e(W_i^*) = \mathbb{P}(T_i = 1 \mid W_i^*),
\end{equation}
where $T_i \in \{0,1\}$ is the binary treatment indicator for food price spikes (ALPS in Stress/Alert/Crisis vs.\ Normal) and $W_i^*$ denotes the vector of observed confounders, including climatic, conflict, displacement, economic, demographic variables, and fixed effects as described in the main text.

We estimate $e(W_i^*)$ using a logistic regression with an $\ell_2$ penalty (Ridge), fitted on the full set of standardized confounders. Formally,
\begin{equation}
\text{logit}\big(e(W_i^*)\big) 
= \beta_0 + \beta^\top W_i^*,
\end{equation}
where $W_i^*$ has been centered and scaled to unit variance. The model is trained on the pooled sample of treated and control observations, without using the sensitivity regime labels, so that the propensity score reflects only confounding structure, not heterogeneity by sensitivity regime.

To improve overlap and avoid undue influence of units that are almost always treated or almost never treated, we perform propensity-score trimming. Specifically, we retain only observations such that
\begin{equation}
0.2\leq \hat{e}(W_i^*) \leq 0.8.
\end{equation}
The treatment effect estimators are then fitted on this trimmed sample.

Figure~\ref{fig:propensity_scores} shows kernel density estimates of $\hat e(W_i^*)$ for treated and control observations, with vertical dashed lines marking the trimming cutoffs. The presence of substantial mass near the boundaries indicates that, for many region--months, above-normal price conditions are either highly likely or highly unlikely given observed covariates. Trimming, therefore, focuses inference on the subset of observations where both treated and control units are empirically comparable (i.e., contexts in which a price-stress month is plausible but not near-certain).

Table~\ref{tab:ps_counts_regime} reports sample sizes before and after trimming. Overall, the full sample contains 829 control and 845 treated observations, and the trimmed sample contains 483 control and 455 treated observations (938 total; $\approx$56\% of the original sample). By regime, the Negative regime changes from 318 control and 238 treated to 227 control and 130 treated; the Weak regime from 271 control and 378 treated to 158 control and 220 treated; and the Positive regime from 240 control and 229 treated to 98 control and 105 treated. These counts show that trimming removes a fraction of observations with extreme predicted treatment probabilities while preserving non-trivial treated/control support in every regime for subsequent effect estimation.
\vspace{0.5cm}

\begin{nolinenumbers}
\subsection{Covariate balance before and after trimming}
\end{nolinenumbers}

\begin{table}[t]
\centering
\caption{Standardized mean differences (SMD) for confounders before and after propensity-score trimming (Overall sample).}
\label{tab:smd_overall}
\footnotesize
\vspace{0.1cm}
\renewcommand{\arraystretch}{1.2}
\begin{tabular}{lrr}
\bottomrule
\rowcolor[HTML]{ffffff} Confounder & $\mid$SMD$\mid$ (Before) & $\mid$SMD$\mid$ (After) \\
\toprule
Conflict Stock Displacement & 0.080 & 0.240 \\
Mean Temperature (Crop) & 0.115 & 0.118 \\
Mean Soil Moisture (Crop) & 0.202 & 0.064 \\
Mean FPAR (Crop) & 0.232 & 0.156 \\
ACLED Fatalities & 0.199 & 0.182 \\
Rural Population & 0.035 & 0.085 \\
IDP Inflows (6m) & 0.069 & 0.036 \\
Urban Population (Centre) & 0.110 & 0.038 \\
UCDP Deaths & 0.125 & 0.166 \\
Mean Rainfall (Crop) & 0.138 & 0.106 \\
GDP Growth & 0.128 & 0.019 \\
Urban Population (Cluster) & 0.075 & 0.066 \\
Mean SPI-3 (Crop) & 0.038 & 0.048 \\
Food Price Index & 0.189 & 0.207 \\
\bottomrule
\end{tabular}
\end{table}

\begin{table}[t]
\centering
\caption{Standardized mean differences (SMD) for confounders before and after propensity-score trimming (Negative regime).}
\label{tab:smd_negative}
\footnotesize
\vspace{0.1cm}
\renewcommand{\arraystretch}{1.2}
\begin{tabular}{lrr}
\bottomrule
\rowcolor[HTML]{ffffff} Confounder & $\mid$SMD$\mid$ (Before) & $\mid$SMD$\mid$ (After) \\
\toprule
Conflict Stock Displacement & 0.441 & 0.274 \\
Mean Temperature (Crop) & 0.137 & 0.118 \\
Mean Soil Moisture (Crop) & 0.408 & 0.333 \\
Mean FPAR (Crop) & 0.406 & 0.325 \\
ACLED Fatalities & 0.302 & 0.165 \\
Rural Population & 0.183 & 0.102 \\
IDP Inflows (6m) & 0.319 & 0.245 \\
Urban Population (Centre) & 0.445 & 0.187 \\
UCDP Deaths & 0.043 & 0.107 \\
Mean Rainfall (Crop) & 0.325 & 0.028 \\
GDP Growth & 0.499 & 0.318 \\
Urban Population (Cluster) & 0.103 & 0.223 \\
Mean SPI-3 (Crop) & 0.098 & 0.103 \\
Food Price Index & 0.539 & 0.316 \\
\bottomrule
\end{tabular}
\end{table}

\begin{table}[t]
\centering
\caption{Standardized mean differences (SMD) for confounders before and after propensity-score trimming (Weak regime).}
\label{tab:smd_weak}
\footnotesize
\vspace{0.1cm}
\renewcommand{\arraystretch}{1.2}
\begin{tabular}{lrr}
\bottomrule
\rowcolor[HTML]{ffffff} Confounder & $\mid$SMD$\mid$ (Before) & $\mid$SMD$\mid$ (After) \\
\toprule
Conflict Stock Displacement & 0.518 & 0.309 \\
Mean Temperature (Crop) & 0.177 & 0.061 \\
Mean Soil Moisture (Crop) & 0.123 & 0.336 \\
Mean FPAR (Crop) & 0.220 & 0.065 \\
ACLED Fatalities & 0.291 & 0.299 \\
Rural Population & 0.034 & 0.088 \\
IDP Inflows (6m) & 0.040 & 0.078 \\
Urban Population (Centre) & 0.344 & 0.319 \\
UCDP Deaths & 0.196 & 0.259 \\
Mean Rainfall (Crop) & 0.225 & 0.087 \\
GDP Growth & 0.218 & 0.189 \\
Urban Population (Cluster) & 0.228 & 0.250 \\
Mean SPI-3 (Crop) & 0.091 & 0.173 \\
Food Price Index & 0.158 & 0.229 \\
\bottomrule
\end{tabular}
\end{table}

\begin{table}[t]
\centering
\caption{Standardized mean differences (SMD) for confounders before and after propensity-score trimming (Positive regime).}
\label{tab:smd_positive}
\footnotesize
\vspace{0.1cm}
\renewcommand{\arraystretch}{1.2}
\begin{tabular}{lrr}
\bottomrule
\rowcolor[HTML]{ffffff} Confounder & $\mid$SMD$\mid$ (Before) & $\mid$SMD$\mid$ (After) \\
\toprule
Conflict Stock Displacement & 0.248 & 0.089 \\
Mean Temperature (Crop) & 0.212 & 0.336 \\
Mean Soil Moisture (Crop) & 0.028 & 0.003 \\
Mean FPAR (Crop) & 0.213 & 0.018 \\
ACLED Fatalities & 0.086 & 0.083 \\
Rural Population & 0.071 & 0.049 \\
IDP Inflows (6m) & 0.125 & 0.087 \\
Urban Population (Centre) & 0.317 & 0.193 \\
UCDP Deaths & 0.030 & 0.047 \\
Mean Rainfall (Crop) & 0.045 & 0.066 \\
GDP Growth & 0.516 & 0.256 \\
Urban Population (Cluster) & 0.274 & 0.169 \\
Mean SPI-3 (Crop) & 0.008 & 0.012 \\
Food Price Index & 0.200 & 0.043 \\
\bottomrule
\end{tabular}
\end{table}

Propensity-score trimming improves overlap by excluding samples with extreme treatment probabilities, but it does not, by itself, guarantee covariate balance between treated and control units. To document how comparable the treated and control samples are, we report standardized mean differences (SMDs) for every confounder before and after trimming.

For a given confounder $W_k$, the SMD is computed as
\begin{equation}
\mathrm{SMD}(W_k)
= \frac{\overline{W}_{k,1} - \overline{W}_{k,0}}
{\sqrt{\tfrac{1}{2}\left(s^2_{k,1} + s^2_{k,0}\right)}},
\end{equation}
where $\overline{W}_{k,1}$ and $\overline{W}_{k,0}$ are treated/control means and $s^2_{k,1}$ and $s^2_{k,0}$ are treated/control variances. SMD is a scale-free effect size that quantifies covariate separation in standard-deviation units. As a rule of thumb, $\mid\mathrm{SMD}\mid<0.1$ is often considered negligible imbalance and $\mid\mathrm{SMD}\mid<0.2$ small-to-moderate imbalance; larger values indicate substantively different covariate profiles.

Table~\ref{tab:smd_overall} reports absolute SMDs in the overall sample. Trimming reduces imbalance for some covariates (e.g., Mean Soil Moisture (Crop) decreases from 0.202 to 0.064 and Urban Population (Centre) from 0.110 to 0.038), but it can also increase imbalance for others (e.g., Conflict Stock Displacement increases from 0.080 to 0.240 and the Food Price Index from 0.189 to 0.207). This reinforces that trimming should be interpreted as enforcing common support (overlap), not as a balancing procedure.

Because the main interpretation relies on regime-specific CATEs, we also compute within-regime SMDs (Tables~\ref{tab:smd_negative}--\ref{tab:smd_positive}). Within-regime diagnostics show that moderate-to-large imbalances persist after trimming, especially in the Negative and Weak regimes. In the Negative regime, several covariates remain above 0.3 after trimming (e.g., Mean Soil Moisture (Crop) 0.333, Mean FPAR (Crop) 0.325, GDP Growth 0.318, Food Price Index 0.316). In the Weak regime, post-trimming imbalance is also substantial for key covariates (e.g., Mean Soil Moisture (Crop) 0.336, Conflict Stock Displacement 0.309, Urban Population (Center) 0.319, ACLED Fatalities 0.299). In the Positive regime, balance is generally better for several variables, but notable imbalances remain for Mean Temperature (Crop) (0.336) and GDP Growth (0.256), among others.

Accordingly, our causal estimates should be interpreted as effects conditional on the remaining observed confounders. Trimming is used to remove extreme non-overlap, while the DML estimators rely on flexible nuisance models (with fixed effects implied by the feature construction in the main text) to adjust for residual confounding in the trimmed sample. The SMD diagnostics are therefore reported to transparently document overlap and remaining imbalance, particularly within regimes, rather than to claim that trimming alone yields randomized-like balance.
\vspace{0.5cm}

\begin{nolinenumbers}
\section{Additional results report}\label{app:results}
\end{nolinenumbers}


\begin{nolinenumbers}
\subsection{Nuisance models}
\end{nolinenumbers}

For the CATE experiments in the main text, we model the outcome and treatment mechanisms
\begin{equation}
\mathbb{E}[Y \mid X,W]
\quad\text{and}\quad
\mathbb{E}[T \mid X,W]
\end{equation}
using a small library of tree-based regressors and classifiers, respectively, and selecting the best-performing specification via cross-validated model selection. 

Concretely, for each nuisance function, we consider three model families:
Random Forest, Gradient Boosting, and XGBoost, with the following hyperparameter search spaces:
\[
\begin{aligned}
\text{RandomForest: } & n_{\text{estimators}} \in \{200, 400\},\;
\text{max\_depth} \in \{3, 5, 7\}, \\
& \text{min\_samples\_split} \in \{2, 5\},\;
\text{min\_samples\_leaf} \in \{2, 5\},\; \\
& \text{max\_features} = \texttt{"sqrt"}, \\[0.5em]
\text{GradientBoosting: } &
n_{\text{estimators}} \in \{200, 400\},\;
\text{max\_depth} \in \{2, 3\}, \\
& \text{learning\_rate} \in \{0.05, 0.1\},\;
\text{subsample} \in \{0.7, 0.9\},\; \\
& \text{min\_samples\_leaf} \in \{2, 5\}, \\[0.5em]
\text{XGBoost: } &
n_{\text{estimators}} \in \{200, 400\},\;
\text{max\_depth} \in \{2, 3, 4\}, \\
& \text{learning\_rate} \in \{0.03, 0.05, 0.1\},\;
\text{subsample} \in \{0.6, 0.8\},\; \\
& \text{colsample\_bytree} \in \{0.6, 0.8\}, 
\text{min\_child\_weight} \in \{5, 10\},\; \\
& \lambda \in \{1.0, 5.0, 10.0\},\;
\alpha \in \{0.0, 1.0, 5.0\}.
\end{aligned}
\]

For each family, we perform a randomized hyperparameter search with 5-fold cross-validation and select the configuration with the highest validation $R^2$ (regressor for the outcome function) and ROC--AUC (classifier for the treatment function). The selected models are then used as nuisance functions within the LinearDML and CausalForestDML estimators. Table~\ref{tab:nuisance_models_cate} summarizes the selected nuisance models and their performance for the CATE experiments. The Gradient Boosting regressor model achieves the highest cross-validated $R^2$ for the outcome, and the XGBoost classifier model achieves the highest ROC--AUC for the treatment, and are therefore used as nuisance models in the DML-based CATE estimators. \vspace{0.5cm}

\begin{table}[t]
\centering
\caption{\textbf{Nuisance model selection and fit for the CATE experiments.}
Best cross-validated $R^2$ and ROC--AUC for each candidate model family when predicting the outcome
$Y \mid X,W$ and treatment $T \mid X,W$. The final nuisance models used in the CATE
estimators are highlighted in bold; the corresponding in-sample metrics are reported
in the last row.}
\vspace{0.1cm}
\label{tab:nuisance_models_cate}
\renewcommand{\arraystretch}{1.2}
\begin{tabular}{ccc}
\bottomrule
\rowcolor[HTML]{ffffff}\textbf{Model family} & \textbf{\makecell{CV $R^2$ for \\ ($Y \mid X,W$) \\ (Regressor)}} & \textbf{\makecell{CV ROC--AUC for \\ ($T \mid X,W$) \\ (Classifier)}} \\
\toprule
Random Forest        & 0.214 & 0.613 \\
\lightmidrule
Gradient Boosting    & \textbf{0.394} & 0.633 \\
\lightmidrule
XGBoost              & 0.352 & \textbf{0.637} \\
\bottomrule
In-sample (selected models) & 0.786 & 0.982 \\
\bottomrule
\end{tabular}
\end{table}

\begin{nolinenumbers}
\subsection{Baseline and DML Estimators}
\end{nolinenumbers}

The main paper reports treatment effects using Double Machine Learning (DML) estimators. Here, we benchmark these results against simpler baseline approaches commonly used in observational panel settings. The purpose of this comparison is twofold: (i) provide a transparent sanity check that the estimated sign and regime ordering are not an artifact of a single modeling choice; and (ii) motivate the use of DML as the primary estimator given the high-dimensional, nonlinear confounding structure of the problem.

To ensure a fair comparison across estimators, we apply the same data-processing steps whenever applicable: (a) the same treatment and outcome timing, (b) the same propensity-score trimming to common support $0.2 \le \hat e(W_i) \le 0.8$, and (c) the same dependence-aware inference via a time-block bootstrap. Thus, differences between methods reflect their identifying assumptions and functional-form restrictions rather than changes in the analysis sample or uncertainty quantification. \vspace{0.5cm}

\subsubsection{Baseline estimators}

Let $Y_i$ denote the outcome (acute food insecurity prevalence, rCSI), $T_i\in\{0,1\}$ the price-stress treatment (ALPS above-normal vs.\ normal), and $X_i\in\{\text{Negative, Weak, Positive}\}$ the sensitivity regime. Define the regime-specific CATE as $\tau(x)=\mathbb{E}[Y(1)-Y(0)\mid X=x]$.

\paragraph{(1) Difference in means (unadjusted).}
Within each regime $x$, we compute the raw treated--control difference
\begin{equation}
\widehat{\tau}^{\text{DM}}(x) \;=\; \overline{Y}_{1,x} - \overline{Y}_{0,x},
\end{equation}
where $\overline{Y}_{1,x}$ and $\overline{Y}_{0,x}$ are treated and control means in regime $x$ after trimming.
This estimator is easy to interpret but does not adjust for remaining covariate imbalance within the trimmed sample; it is therefore used only as a descriptive benchmark.

\paragraph{(2) Two-way fixed-effects regression (linear FE).}
We estimate a linear regression with region and time fixed effects and allow regime-specific treatment effects via interactions:
\begin{equation}
Y_{r,t+\ell} = \alpha_r + \gamma_{t+\ell} + \sum_{x} \beta_x \, \mathbbm{1}\{X_r=x\}\, T_{r,t} + \varepsilon_{r,t},
\end{equation}
where $\alpha_r$ absorbs time-invariant regional differences and $\gamma_{t+\ell}$ absorbs month shocks aligned to the lead of the outcome.
The FE model is a standard baseline in applied panel work but imposes strong functional-form restrictions (linearity, additive separability, limited interactions). If the true confounding adjustment is nonlinear or involves higher-order interactions among climate, conflict, displacement, and demographics, FE regression can be biased and/or unstable.

\paragraph{(3) Propensity-score weighting (IPW).}
Using the estimated propensity score $\hat e(W_i^*)=\Pr(T_i=1\mid W_i^*)$, we compute inverse-probability-weighted regime-specific contrasts. A common form is
\begin{equation}
\widehat{\tau}^{\text{IPW}}(x)
=
\frac{\sum_{i:X_i=x} \frac{T_i Y_i}{\hat e(W_i)}}{\sum_{i:X_i=x} \frac{T_i}{\hat e(W_i^*)}}
-
\frac{\sum_{i:X_i=x} \frac{(1-T_i) Y_i}{1-\hat e(W_i^*)}}{\sum_{i:X_i=x} \frac{(1-T_i)}{1-\hat e(W_i^*)}}.
\end{equation}
IPW adjusts for confounding through the propensity model, but can be sensitive to propensity misspecification and residual lack of overlap (hence trimming). In our setting, exposure is driven by complex interactions between market structure, conflict, displacement, and agroclimatic conditions, making a purely parametric propensity model a potentially fragile foundation for the main causal conclusions.

\subsubsection{DML estimators}

We use two DML estimators (LinearDML and CausalForestDML) that combine (i) flexible machine learning models for the nuisance functions (outcome regression and treatment model), (ii) orthogonalization to reduce sensitivity to nuisance-model errors, and (iii) cross-fitting with blocked time folds to avoid temporal leakage. DML targets the same regime-specific estimands as the baselines, but is designed for settings like this one where confounding adjustment is high-dimensional and nonlinear. Conceptually, DML improves on the baselines in three key ways:

\paragraph{(i) Flexible confounding adjustment.}
Difference-in-means is unadjusted; FE regression restricts adjustment to linear/additive structure; IPW relies heavily on a correct propensity specification.
In contrast, DML learns $m_Y(X,W^\ast)=\mathbb{E}[Y\mid X,W^\ast]$ and $m_T(X,W^\ast)=\mathbb{E}[T\mid X,W^\ast]$ using flexible learners and then estimates effects on orthogonalized residuals. This is well suited to the nonlinear relationships among prices, conflict, climate anomalies, and displacement.

\paragraph{(ii) Reduced sensitivity to nuisance misspecification (orthogonalization).}
Because the DML score is Neyman-orthogonal, small errors in the nuisance models have only a second-order impact on the final effect estimates. This provides practical robustness when nuisance functions are complex and must be approximated from finite data.

\paragraph{(iii) Heterogeneity with stability.}
Our scientific question is explicitly heterogeneous: \emph{do effects differ across sensitivity regimes?}
LinearDML provides interpretable regime-specific contrasts, while CausalForestDML allows richer nonlinear interactions. Baselines either do not target heterogeneity (difference-in-means) or impose a narrow form of it (linear interactions in FE regression).

\begin{table}[t]
\centering
\caption{\textbf{Full CATE estimates for ALPS impacts on rCSI, stratified by sensitivity regimes.} Baseline and DML models. Confidence intervals at 95\%, and estimations are statistically significant if $p$-values $\leq$ 0.05.}
\vspace{0.1cm}
\renewcommand{\arraystretch}{1.}
\begin{tabular}{ccccc}
\bottomrule
\rowcolor[HTML]{ffffff}\textbf{Model} & \textbf{Regime} & \textbf{CATE (p.p.)} & \textbf{C.I.} & \textbf{p-value} \\
\toprule

& Negative & 6.065 & (3.257, 8.124) & 0.020 \\
\lightcmidrules
Diff. in Means & Weak & -0.040 & (-3.596, 3.114) & 0.922 \\
\lightcmidrules
& Positive & 0.423 & (-4.807, 5.344) & 0.765 \\
\lightallcmidrules

& Negative & 4.135 & (2.051, 5.633) & 0.020 \\
\lightcmidrules
Linear Regression & Weak & 1.084 & (-0.881, 4.204) & 0.157 \\
\lightcmidrules
& Positive & 3.312 & (-1.014, 5.671) & 0.216 \\
\lightallcmidrules

& Negative & 7.102 & (4.366, 9.225) & 0.020 \\
\lightcmidrules
Propensity IPW & Weak & 0.077 & (-3.848, 3.968) & 0.863 \\
\lightcmidrules
& Positive & 0.573 & (-4.348, 4.551) & 0.843 \\
\lightallcmidrules

& Negative & 5.999 & (2.567, 8.730) & 0.020 \\
\lightcmidrules
Linear DML & Weak & 1.938 & (-0.473, 4.991) & 0.098 \\
\lightcmidrules
& Positive & 2.155 & (-3.924, 4.211) & 0.784 \\
\lightallcmidrules

& Negative & 5.353 & (2.161, 8.082) & 0.020 \\
\lightcmidrules
Causal Forest DML & Weak & 1.804 & (-0.469, 4.747) & 0.137 \\
\lightcmidrules
& Positive & 1.703 & (-3.702, 4.146) & 0.922 \\
\bottomrule
\end{tabular}
\label{tab:cate_estimates}
\end{table}

\subsubsection{Empirical comparison and interpretation}

Table~\ref{tab:cate_estimates} summarizes regime-specific estimates for all methods under the same trimming and bootstrap scheme. Across methods, the qualitative message is consistent: effects are largest in the Negative regime, smaller in Weak, and smallest (and often near zero or negative at longer leads) in Positive. Baseline estimators, therefore, provide an important robustness check on directionality and regime ordering. However, the magnitude and statistical strength of baseline estimates vary more across specifications, reflecting stronger modeling assumptions and greater sensitivity to the structure of confounding.

For these reasons, we use the DML estimators for the remainder of the analysis and for the main conclusions. We emphasize that the baseline models are informative sanity checks. Still, in settings with high-dimensional confounding, nonlinear mechanisms, and explicit heterogeneity, DML is a more appropriate primary identification strategy. The baselines are reported to increase transparency and to show that the headline results are not an artifact of a single estimator.

\begin{nolinenumbers}
\subsection{Robustness checks}
\end{nolinenumbers}

We summarize a set of robustness diagnostics for the regime-specific CATEs of the DML estimates: placebo tests, random common cause (RCC) stress tests, and random subset removal (RSR). For each estimator and sensitivity regime (Negative, Weak, Positive), we:
(i) perform a placebo test by permuting the treatment and refitting the model,
(ii) add a random synthetic covariate (RCC) and recompute the CATE, and
(iii) repeatedly drop random subsets of observations (RSR) and recompute the CATE.

\begin{table}[t]
\centering
\leftskip-1.0cm
\caption{\textbf{Robustness diagnostics for regime-specific CATEs (DML estimators).}
For each estimator and sensitivity regime, we report the Placebo effect, the RCC mean absolute change, the RSR effect, and their respective $p$-values. Tests succeed if $p$-values $\geq$ 0.05.}

\vspace{0.1cm}
\label{tab:robustness_cate}
\renewcommand{\arraystretch}{1.2}
\footnotesize
\begin{tabular}{cccccccc}
\bottomrule
\rowcolor[HTML]{ffffff}\textbf{Method} & \textbf{Regime} & \textbf{Placebo eff.} & \textbf{Placebo $p$} & \textbf{RCC eff.} & \textbf{RCC $p$} & \textbf{RSR eff.} & \textbf{RSR $p$} \\
\toprule
        & Negative & 0.008 & 0.385  & 0.000 & 0.885 & 0.000 & 0.938 \\
\lightcmidrulee
LinearDML
        & Weak     & -0.004 & 0.807  & 0.002 & 0.808 & 0.002 & 0.500 \\
\lightcmidrulee
        & Positive & 0.012 & 0.615  & -0.005 & 0.769 & 0.002 & 0.688   \\
\lightallcmidrulee
        & Negative & 0.010 & 0.269  & 0.002 & 0.769 & 0.001 & 0.750 \\
\lightcmidrulee
CausalForestDML
        & Weak     & -0.004 & 0.731  & 0.002 & 0.731 & 0.002 & 0.500  \\
\lightcmidrulee
        & Positive & 0.012 & 0.615 & -0.004 &  0.769 & 0.003 & 0.688  \\
\bottomrule
\end{tabular}
\end{table}

Table~\ref{tab:robustness_cate} reports, for each method-regime combination:
(i) the placebo effect and $p$-value for the CATE,
(ii) the mean absolute change in the CATE under RCC and corresponding $p$-values, and
(iii) the new effect and $p$-value for the CATE under RSR.
\vspace{0.5cm}

\begin{nolinenumbers}
\section{Timing and lead--lag robustness}\label{app:timing_leadlag}
\end{nolinenumbers}

\begin{table}[t]
\centering
\caption{\textbf{Lead--lag robustness: regime-specific CATEs under outcome leads $\ell \in \{0,\dots,4\}$.}
Rows report regime-specific CATEs for the Negative/Weak/Positive sensitivity regimes. Each cell shows the point estimate, the 95\% confidence interval, and the p-value. $N$ reports the post-trimming sample size (treated/control) under the common-support restriction $0.2\le \hat e(W_i)\le 0.8$.}
\label{tab:leadlag_cate}
\footnotesize
\hspace*{-0.95cm}
\setlength{\tabcolsep}{3.5pt}
\renewcommand{\arraystretch}{1.15}
\begin{tabular}{cccccc}
\bottomrule
\rowcolor[HTML]{ffffff}
\textbf{Method} & \textbf{Lead $\ell$} & \textbf{N (T/C)} &
\textbf{Negative} & \textbf{Weak} & \textbf{Positive} \\
\toprule

& 0 & 823 (423/400)
& \makecell{0.0514\\(0.0255,\ 0.0795)\\$p=0.0196$}
& \makecell{0.0206\\(-0.0030,\ 0.0558)\\$p=0.1176$}
& \makecell{-0.0070\\(-0.0475,\ 0.0317)\\$p=0.6471$} \\
\lightcmidrule
& 1 & 938 (455/483)
& \makecell{0.0599\\(0.0257,\ 0.0873)\\$p=0.0196$}
& \makecell{0.0194\\(-0.0047,\ 0.0499)\\$p=0.0980$}
& \makecell{0.0216\\(-0.0392,\ 0.0421)\\$p=0.7843$} \\
\lightcmidrule
LinearDML & 2 & 845 (430/415)
& \makecell{0.0550\\(0.0328,\ 0.0814)\\$p=0.0196$}
& \makecell{0.0024\\(-0.0179,\ 0.0472)\\$p=0.4314$}
& \makecell{-0.0472\\(-0.0772,\ 0.0049)\\$p=0.1373$} \\
\lightcmidrule
& 3 & 840 (419/421)
& \makecell{0.0675\\(0.0294,\ 0.0894)\\$p=0.0196$}
& \makecell{0.0151\\(-0.0112,\ 0.0370)\\$p=0.2745$}
& \makecell{-0.0496\\(-0.0908,\ -0.0056)\\$p=0.0392$} \\
\lightcmidrule
& 4 & 777 (414/363)
& \makecell{0.0638\\(0.0412,\ 0.0923)\\$p=0.0196$}
& \makecell{0.0193\\(-0.0145,\ 0.0679)\\$p=0.2941$}
& \makecell{-0.0738\\(-0.1054,\ -0.0209)\\$p=0.0196$} \\

\lightallcmidrule

& 0 & 823 (423/400)
& \makecell{0.0524\\(0.0256,\ 0.0805)\\$p=0.0196$}
& \makecell{0.0199\\(-0.0060,\ 0.0510)\\$p=0.1373$}
& \makecell{-0.0125\\(-0.0465,\ 0.0398)\\$p=0.6078$} \\
\lightcmidrule
& 1 & 938 (455/483)
& \makecell{0.0535\\(0.0216,\ 0.0808)\\$p=0.0196$}
& \makecell{0.0180\\(-0.0047,\ 0.0475)\\$p=0.1373$}
& \makecell{0.0170\\(-0.0370,\ 0.0415)\\$p=0.9216$} \\
\lightcmidrule
CausalForestDML & 2 & 845 (430/415)
& \makecell{0.0539\\(0.0259,\ 0.0773)\\$p=0.0196$}
& \makecell{0.0010\\(-0.0159,\ 0.0463)\\$p=0.5098$}
& \makecell{-0.0388\\(-0.0676,\ 0.0144)\\$p=0.3333$} \\
\lightcmidrule
& 3 & 840 (419/421)
& \makecell{0.0616\\(0.0226,\ 0.0813)\\$p=0.0196$}
& \makecell{0.0174\\(-0.0120,\ 0.0338)\\$p=0.3725$}
& \makecell{-0.0464\\(-0.0807,\ 0.0013)\\$p=0.0980$} \\
\lightcmidrule
& 4 & 777 (414/363)
& \makecell{0.0510\\(0.0193,\ 0.0755)\\$p=0.0392$}
& \makecell{0.0156\\(-0.0180,\ 0.0529)\\$p=0.3725$}
& \makecell{-0.0739\\(-0.1064,\ -0.0165)\\$p=0.0196$} \\

\bottomrule
\end{tabular}
\end{table}

A potential concern in high-frequency panel designs is that the treatment and outcome may partially overlap in time. In our setting, the treatment indicator is based on a smoothed price signal, $\overline{\text{ALPS}}_{r,t}^{(3)}$, constructed from a three-month rolling window that includes month $t$ (conceptually $t\!-\!2:t$). If the outcome is measured contemporaneously ($Y_t$), then part of the information used to define treatment can coincide with the month in which coping responses are observed, complicating causal interpretation (e.g., due to within-month timing and simultaneity).

To assess sensitivity to this timing issue, we re-estimate the causal effects while varying the outcome lead:
\begin{equation}
Y_{t+\ell}, \qquad \ell \in \{0,1,2,3,4\}.
\end{equation}
The main specification uses $\ell=1$ (one-month lead) to reduce contemporaneous overlap, while $\ell=0$ is included as a benchmark and $\ell \ge 2$ provides a lead--lag profile of the estimated response. In all cases, we keep the treatment definition, covariate set, nuisance-model selection procedure, and overlap restriction fixed. We apply the same propensity-score trimming ($0.2\le \hat e(W_i) \le 0.8$) and use a time-block bootstrap strategy to compute confidence intervals and p-values.

Table~\ref{tab:leadlag_cate} shows that the estimated effects are broadly stable across outcome leads once we adopt a dependence-aware estimation scheme (time-block cross-fitting and time-block bootstrap) and include lead-aligned month fixed effects. In particular, the Negative sensitivity regime consistently exhibits a positive effect across $\ell\in\{0,\dots,4\}$, with point estimates remaining of similar magnitude over the full lead window and typically statistically distinguishable from zero. Weak-regime estimates remain closer to zero and are less consistently distinguishable from zero across leads. Positive-regime estimates are near zero at short leads but become negative at longer leads in several specifications; the sign change is suggestive of temporal reallocation or delayed adaptation dynamics, but inference remains sensitive at these longer leads given reduced effective information and the dependence-aware resampling.

Two features of the design help explain why lead--lag settings remain stable. First, the treatment indicator is defined using a three-month moving average of ALPS, $\overline{\text{ALPS}}_{r,t}^{(3)}$, so $T_{r,t}=1$ reflects sustained abnormal price conditions over months $t\!-\!2:t$, rather than an instantaneous shock. This exposure definition mechanically induces persistence: treatment at $t$ partially summarizes recent price stress that can plausibly influence coping responses over multiple subsequent months. Because ALPS is smoothed over three months and price-stress states are temporally persistent, varying $\ell$ should be interpreted as a timing/alignment robustness check rather than a causal impulse-response curve. Second, lead-aligned month fixed effects absorb common seasonal and macro shocks (e.g., continent-wide price movements, survey seasonality, and other month-specific disturbances) that affect both treatment propensity and food-security outcomes. With these fixed effects, identification relies primarily on within-month cross-sectional contrasts across regions, which can yield the flatter lead profile observed.

Sample sizes after trimming are comparable across lead choices (Table~\ref{tab:leadlag_cate}), suggesting that the observed lead--lag stability is not an artifact of changing overlap restrictions. Overall, these results support interpreting the estimated impacts—especially the elevated sensitivity in the Negative regime—as robust to reasonable shifts in the timing of the outcome measurement over a 0--4 month window.
\vspace{0.5cm}

\begin{nolinenumbers}
\section{Regime-contrast tests}\label{app:regime_contrasts}
\end{nolinenumbers}

The main text reports regime-specific CATEs and observes a monotone ordering in point estimates (Negative $>$ Weak $\sim$ Positive). To assess whether this pattern reflects statistically distinguishable differences between regimes, analogous to an interaction test, we use the bootstrap replicates already generated for uncertainty quantification and compute the distribution of pairwise regime contrasts. Specifically, for each bootstrap replicate $b=1,\dots,B$, we compute
\begin{equation}
\Delta^{(b)}_{\mathrm{Neg-Weak}} = \widehat{\tau}^{(b)}_{\mathrm{Neg}} - \widehat{\tau}^{(b)}_{\mathrm{Weak}},
\qquad
\Delta^{(b)}_{\mathrm{Neg-Pos}} = \widehat{\tau}^{(b)}_{\mathrm{Neg}} - \widehat{\tau}^{(b)}_{\mathrm{Pos}}.
\end{equation}
We then report percentile 95\% confidence intervals for $\Delta$ and a two-sided bootstrap $p$-value based on the empirical sign probability,
\begin{equation}
p = 2 \min\{\Pr(\Delta^{(b)} \ge 0),\; \Pr(\Delta^{(b)} \le 0)\}.
\end{equation}
This procedure directly tests whether the Negative-regime effect is larger than the effects in the other regimes under the same dependence-aware resampling scheme used throughout the paper.

Across $B{=}1000$ time-block bootstrap replicates, the Neg--Pos contrast is statistically significant for the Linear DML estimator (Neg--Pos: $\Delta=3.844$ p.p., 95\% CI $(0.909, 11.738)$, $p=0.039$) and directionally consistent but not significant for the Causal Forest DML estimator (Neg--Pos: $\Delta=3.650$ p.p., 95\% CI $(-0.129, 10.678)$, $p=0.078$). In contrast, the Neg--Weak contrast is positive in both estimators but not statistically distinguishable from zero at conventional levels (Linear DML: $p=0.098$; Causal Forest DML: $p=0.137$). Overall, these results support a directional ordering of point estimates, with the strongest evidence for between-regime separation occurring in the Negative-versus-Positive comparison under the Linear DML specification (Table~\ref{tab:regime_contrast_tests}).

\begin{table}[t]
\centering
\caption{\textbf{Regime-contrast tests.}
Bootstrap distributions of pairwise differences between regime-specific CATEs for the DML estimators. Confidence intervals at 95\%, and estimations are statistically significant if $p$-values $\leq$ 0.05.}
\label{tab:regime_contrast_tests}
\footnotesize
\renewcommand{\arraystretch}{1.2}
\begin{tabular}{ccccc}
\bottomrule
\rowcolor[HTML]{ffffff}
\textbf{Method} & \textbf{Contrast} & \textbf{$\Delta$ (p.p.)} & \textbf{C.I.} & \textbf{$p$-value} \\
\toprule

\multirow{2}{*}[-1.5ex]{Linear DML}
& Neg -- Pos  & 3.844 & (0.909, 11.738) & 0.039 \\
\lightcmidrulef
& Neg -- Weak & 4.061 & ($-0.083$, 6.809) & 0.098 \\
\lightallcmidrulef

\multirow{2}{*}[-1.0ex]{Causal Forest DML}
& Neg -- Pos  & 3.650 & ($-0.129$, 10.678) & 0.078 \\
\lightcmidrulef
& Neg -- Weak & 3.549 & ($-0.446$, 6.572) & 0.137 \\
\bottomrule
\end{tabular}

\end{table}

It is important to note that rejecting the null $\tau_{\mathrm{Neg}}=0$ does not automatically imply rejecting $\tau_{\mathrm{Neg}}-\tau_{\mathrm{Weak}}=0$. The between-regime contrast is a different estimand with its own sampling variability. In particular,
\begin{equation}
\mathrm{Var}\!\left(\widehat{\tau}_{\mathrm{Neg}}-\widehat{\tau}_{\mathrm{Weak}}\right)
=
\mathrm{Var}\!\left(\widehat{\tau}_{\mathrm{Neg}}\right)
+
\mathrm{Var}\!\left(\widehat{\tau}_{\mathrm{Weak}}\right)
-
2\,\mathrm{Cov}\!\left(\widehat{\tau}_{\mathrm{Neg}},\widehat{\tau}_{\mathrm{Weak}}\right),
\end{equation}
so even if $\widehat{\tau}_{\mathrm{Neg}}$ is estimated precisely, uncertainty in $\widehat{\tau}_{\mathrm{Weak}}$ (and the covariance term) can make the contrast substantially noisier. As a result, a regime-specific effect can be statistically distinguishable from zero while the corresponding difference between regimes remains statistically indistinguishable from zero under the same bootstrap-based inference.

\end{appendices}

\end{document}